\pdfoutput=1

\documentclass[11pt]{article}
\usepackage[final]{acl}

\usepackage{times}
\usepackage{latexsym}

\usepackage[T1]{fontenc}
\usepackage[utf8]{inputenc}

\usepackage{microtype}

\usepackage{inconsolata}

\usepackage{graphicx}
\usepackage{booktabs}
\usepackage{multirow}
\usepackage{color}
\usepackage{hyperref}
\usepackage{amsmath}
\usepackage{amsfonts}
\usepackage{amsthm}
\usepackage{scalefnt}
\usepackage{xspace}
\usepackage{authblk}
\usepackage{dcolumn,booktabs}
\usepackage{amsmath}
\usepackage{graphicx}
\usepackage{amsfonts}
\usepackage{multirow}
\usepackage{listings}

\usepackage[most]{tcolorbox}
\tcbset{
  colback=gray!3,
  colframe=black!15,
  boxrule=0.3pt,
  left=1mm,right=1mm,top=0.5mm,bottom=0.5mm,
  enlarge left by=0mm, enlarge right by=0mm
}

\newtcolorbox{PromptBox}[2]{%
  title={\textbf{#1} \hfill \normalfont\footnotesize},
  fonttitle=\bfseries,
}

\theoremstyle{plain}

\def\secref#1{\S\ref{sec:#1}}
\def\seclabel#1{\label{sec:#1}}

\newcommand{\algorithmname}{\textsc{MulTypo}\xspace}

\newcounter{notecounter}

\newcommand{\enoteson}{\long\gdef\enote##1##2{{
\stepcounter{notecounter}
{\large\bf
\hspace{0cm}\arabic{notecounter} $<<<$ ##1: ##2
$>>>$\hspace{1cm}}}}}

\enoteson

\title{Evaluating Robustness of Large Language Models \\ Against Multilingual Typographical Errors}

\author[1,2,*]{\bf{Raoyuan Zhao}}
\author[1,2,*,$^\diamond$]{\bf{Yihong Liu}}
\author[1]{\bf{Lena Altinger}}
\author[1,2]{\\{\bf Hinrich Sch\"utze}}
\author[1,2]{\bf Michael A. Hedderich}

\affil[1]{Center for Information and Language Processing, LMU Munich}
\affil[2]{Munich Center for Machine Learning (MCML)
 \protect\\ \texttt{\{rzhao, yihong, hedderich\}@cis.lmu.de}}

\begin{document}
\maketitle

\def\thefootnote{*}\footnotetext{Equal contribution.}\def\thefootnote{\arabic{footnote}}
\def\thefootnote{$\diamond$}\footnotetext{Corresponding author.}\def\thefootnote{\arabic{footnote}}

\begin{abstract}

Large language models (LLMs) are increasingly deployed in multilingual, real-world applications with user inputs -- naturally introducing \emph{typographical errors} (typos). 
Yet most benchmarks assume clean input, leaving the robustness of LLMs to typos across languages largely underexplored. 
To address this gap, we introduce \textbf{\algorithmname}, a multilingual typo generation algorithm that simulates human-like errors based on language-specific keyboard layouts and typing behavior. 
We evaluate 18 open-source LLMs across three model families and five downstream tasks spanning language inference, multi-choice question answering, mathematical reasoning, and machine translation tasks.
Our results show that typos consistently degrade performance, particularly in generative tasks and those requiring reasoning -- while the natural language inference task is comparatively more robust. 
Instruction tuning improves clean-input performance but may increase brittleness under noise. 
We also observe language-dependent robustness: high-resource languages are generally more robust than low-resource ones, and translation from English is more robust than translation into English.
Our findings underscore the need for noise-aware training and multilingual robustness evaluation. 
We release a Python package for \algorithmname and make the source code publicly available at \url{https://github.com/cisnlp/multypo}.

\end{abstract}

\section{Introduction}
\begin{figure}[t]
    \centering
    \includegraphics[width=0.95\linewidth]{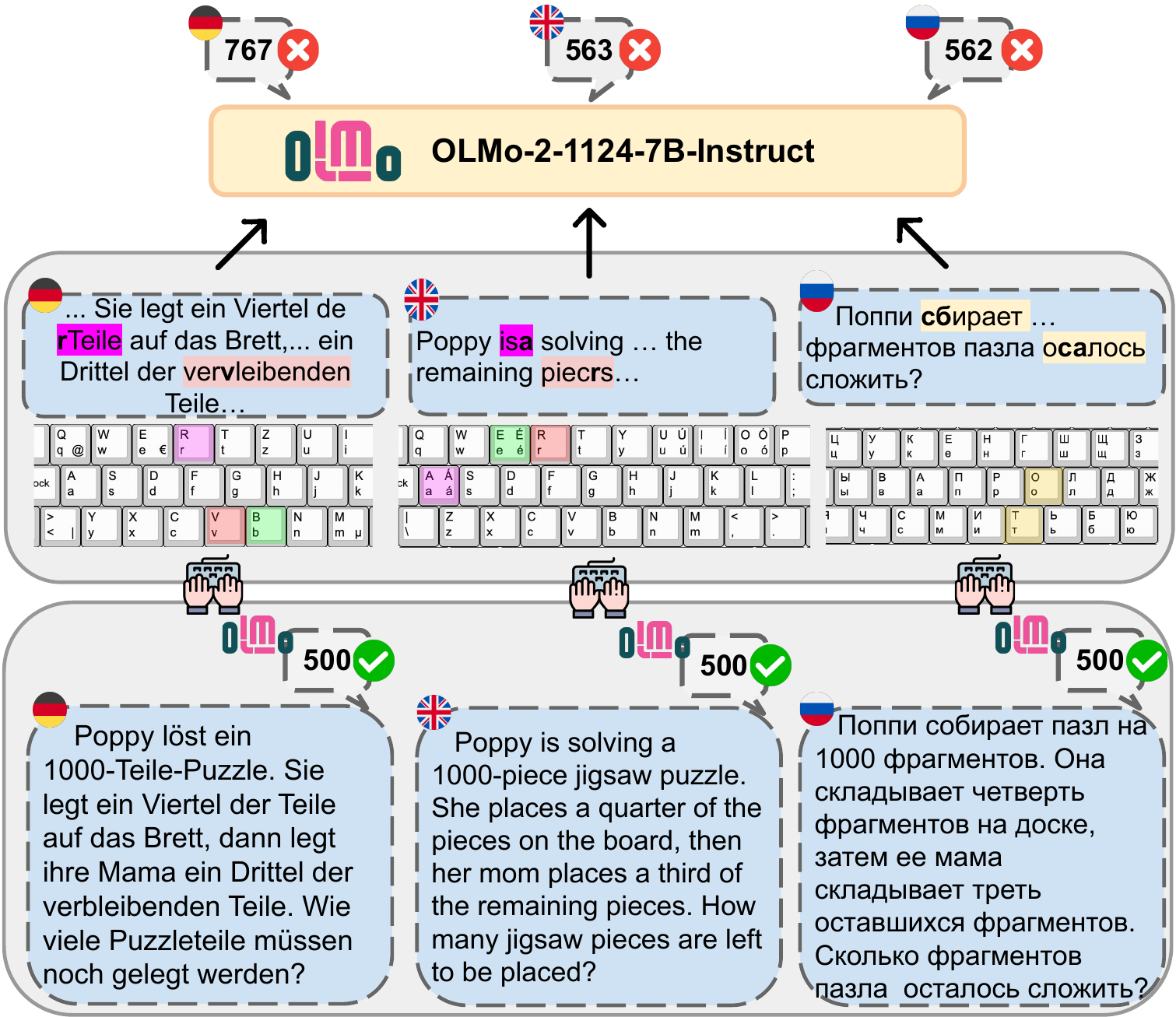}
    \caption{Illustration of the impact of real-world typographical errors. 
    Humans often make typos on language-specific keyboard layouts, and once such errors are introduced, models can fail across languages.
    In this example, the model cannot generate the correct answer ("500") under typos in English, German, and Russian.
    }
    \label{fig:typo-intro}
\end{figure}
LLMs are increasingly deployed in real-world applications such as chatbots, translation tools, and search engines \citep{dam2024completesurveyllmbasedai,naveed2024comprehensiveoverviewlargelanguage,raza2025industrial}, where users input text via keyboards in a wide range of languages.
In such settings, 
\emph{typographical errors} (typos)
are a natural part of user input -- arising from slips, fast typing, or unfamiliarity with keyboard layouts \citep{wengelin2007word,conijn2019typo,shi2025simulating}.
Despite this, most LLM evaluations assume clean, error-free input and report only a single aggregate statistic on a held-out set, thereby overlooking this ubiquitous source of noise \citep{sun2020advbertbertrobustmisspellings,moradi-samwald-2021-evaluating,wang-etal-2024-resilience} and often overestimating real-world performance \citep{ribeiro-etal-2020-beyond,Hedderich2021Label,zhao-etal-2024-syntheval}
Robustness to typos is not just a usability concern; it is essential for ensuring reliable model behavior, maintaining user trust, and delivering consistent downstream performance in practical deployments.

Prior work on robustness evaluation has largely focused on adversarial or synthetic perturbations in an English-centric manner \citep{gao2018black,wang2023robustnesschatgptadversarialoutofdistribution,gan-etal-2024-reasoning,zhu2024promptrobust,zhang2025evaluatingimprovingrobustnesslarge,schmidtova-etal-2026-important}.
As a result, we know little about how robust LLMs are against realistic typos in a multilingual context.
Models can generate wrong answers with simple input textual perturbations across languages, as shown in Figure~\ref{fig:typo-intro}.
Moreover, these approaches often rely on edit-distance heuristics or character-level noise, with little regard for typing behavior (e.g., \emph{10-finger typing convention}) based on language-specific keyboard layouts.

\begin{table}
\centering
    \setlength{\belowcaptionskip}{-0.3cm}
\footnotesize
\begin{tabular}{ll}
\toprule
\textbf{Error} & \textbf{Example Sentence} \\
\hline 
None & Colorless green ideas smell furiously. \\
\hline
Replacement & Colorless green idea\textcolor{red}{a} smell furiously. \\
Insertion & Colorless green\textcolor{red}{m} ideas smell furiously. \\
Deletion & \textcolor{red}{Coorless} green ideas smell furiously. \\
Transposition & Colorless green ideas smell furiou\textcolor{red}{ls}y. \\
\bottomrule
\end{tabular}
\caption{Examples of four different typographical errors.}\label{tab:illustrations-of-typos}
\end{table}

To address these gaps, we first introduce \textbf{\algorithmname}, a multilingual typo generation algorithm grounded in empirical modeling of typing behavior.
Unlike prior work, \algorithmname simulates realistic typos based on actual language-specific keyboard layouts and constraints, allowing us to generate noise that better reflects real-world user patterns across languages.
Crucially, we validate typo realism through human evaluation, ensuring our perturbations reflect actual typing behavior.
Relying on \algorithmname, we conduct a comprehensive robustness evaluation of 18 open-source LLMs, spanning three major model families: Gemma, Qwen, and OLMo, using both base
models and their instruction-tuned versions, across diverse downstream tasks. 
Moreover, we assess the model robustness under varying levels of typo corruption, and under both zero- and few-shot prompting.

Our experiments yield several key findings. 
First, typographical errors consistently degrade model performance, particularly on generative tasks and those requiring reasoning (\secref{typo_rate}). 
Second, model size does not guarantee robustness: both small and large models exhibit noticeable performance drops under typos (\secref{model_size}).
Third, instruction tuning improves clean-input performance but may also increase vulnerability under noise (\secref{instruction_tuned}).
Fourth, increasing the number of examples in few-shot settings
does not improve the robustness against typos (\secref{shot_count}).
Lastly, robustness varies substantially across languages and scripts: for instance, translation \emph{from} English tends to be more robust than translation \emph{into} English (\secref{language_sensitive}). 

Our contributions are summarized as follows.
\textbf{(i)} We propose \textbf{\algorithmname}, a multilingual typo generation algorithm that simulates realistic human-like errors grounded in language-specific keyboard layouts and typing patterns, and validate its realism through human evaluation.
\textbf{(ii)} We conduct a comprehensive robustness evaluation suite spanning 18 open-source LLMs from three families (Gemma, Qwen, OLMo), across five downstream tasks.
\textbf{(iii)} We evaluate robustness under both zero-shot and few-shot prompting, and under varying levels of typographical corruption, enabling fine-grained analysis of model behavior.
\textbf{(iv)} We release a Python package for \algorithmname, along with its source code, to facilitate further research on multilingual robustness against typographical errors.

\section{Background and Related Work}

\subsection{Typographical Errors}\seclabel{background_typos}

\emph{Typographical errors} are among the most common forms of natural noise in user-generated text, typically resulting from accidental keystrokes during typing.
Early studies \citep{gardner1992spelling, lisbach2013linguistic} have identified four core types of typos: 
\textbf{replacement}, \textbf{insertion}, \textbf{deletion}, and \textbf{transposition}.
\emph{Replacement} errors occur when the intended key is substituted with another, typically an adjacent key.
\emph{Insertion} errors arise from unintentionally pressing an adjacent key alongside the intended one, while \emph{deletion} errors involve accidentally omitting a character. 
\emph{Transposition} errors, frequently attributed to asynchronous hand movements, swap two adjacent characters, particularly those typed by different hands.
Examples are illustrated in Table~\ref{tab:illustrations-of-typos}.
These four categories have been widely adopted and extended in subsequent work on noise modeling and robustness evaluation \citep{gao2018black, pruthi-etal-2019-combating, zhang-etal-2022-interpreting, gan-etal-2024-reasoning}.
They also form the foundation of our multilingual typo simulation algorithm \algorithmname (cf. \secref{multypo}),\footnote{For comparison, we implement a \emph{naive baseline} that applies the same operations without layout constraints, in line with prior approaches. We show that our typos better resemble human errors (cf. \secref{user_study}) and that models exhibit different robustness to our typos versus the naive ones (cf. \secref{comparison_baseline}).} which extends this line of work by incorporating language-specific keyboard layouts.



\subsection{Related Work}

\begin{figure*}
    \centering
    \setlength{\belowcaptionskip}{-0.2cm}
    \includegraphics[width=0.98\textwidth]{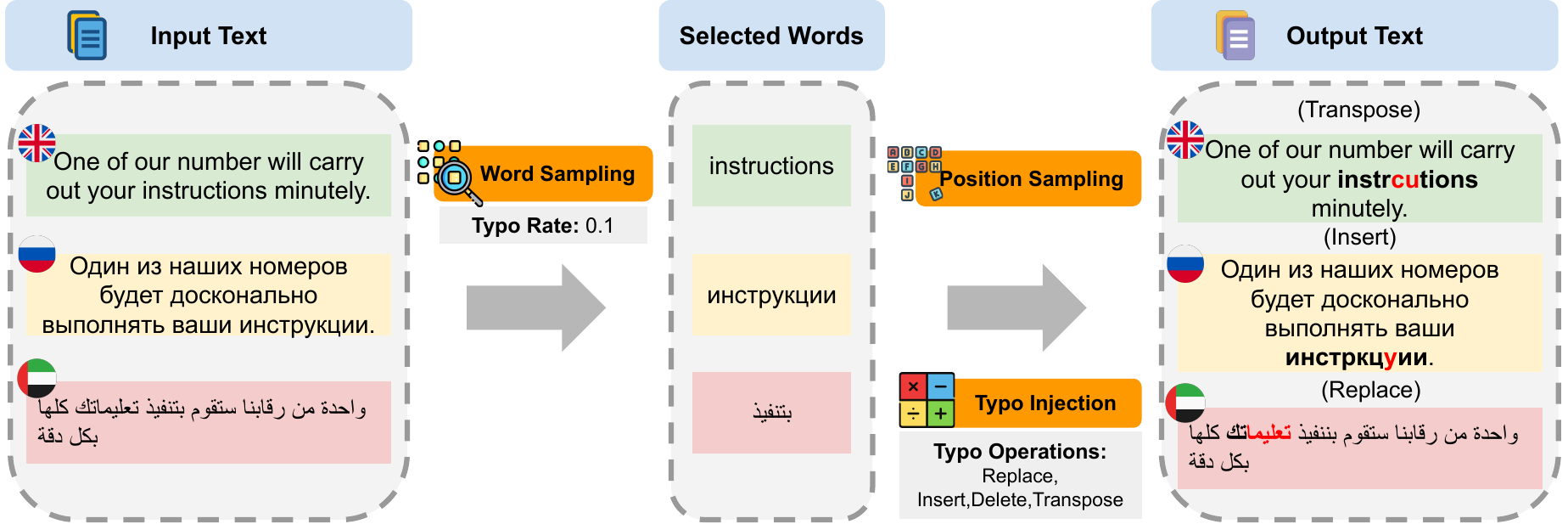}
    \caption{Illustration of the pipeline of \algorithmname: Given an input text and a user-defined typo rate, the algorithm (i) samples words with probability proportional to the square root of the word length, (ii) samples character positions using a position-aware distribution, and (iii) samples one of four typo operations: \emph{replace}, \emph{insert}, \emph{delete}, or \emph{transpose}. Then the algorithm produces a noised text that simulates human-like errors.
    }
    \label{fig:pipeline}
\end{figure*}

\paragraph{Input Text Perturbations}
Recent advances in LLMs have sparked growing interest in evaluating their robustness to noisy or manipulated input. 
A large body of work focuses on \emph{input corruptions}, which aim to degrade model performance by perturbing the input text at various granularities.
These include character-level modifications such as misspellings and typographical errors \citep{gao2018black, li2019textbugger, pruthi-etal-2019-combating, gan-etal-2024-reasoning}, word-level attacks like synonym substitution or word shuffling \citep{garg-ramakrishnan-2020-bae, jin2020bert, moradi-samwald-2021-evaluating, zhou2024mathattack}, and sentence-level perturbations through paraphrasing or irrelevant context insertions \citep{shi2023large, lanham2023measuring, xu2024llm}. 
Even minor punctuation noise can affect LLMs' performance \citep{abedin2025arithmattackevaluatingrobustnessllms}.
However, most of this research is conducted in English-centric settings, leaving it unclear how typos influence LLMs' robustness across different languages -- a gap our work directly addresses.

\paragraph{Multilingual Robustness Evaluation} Another line of work has examined the robustness of multilingual models \citep{cooper-stickland-etal-2023-robustification, aliakbarzadeh2025exploringrobustnessmultilingualllms,okewunmi-etal-2025-evaluating}.
For example, \citet{cooper-stickland-etal-2023-robustification} investigated how real-world noise influences encoder-only models such as XLM-R \citep{conneau-etal-2020-unsupervised} and mBERT \citep{devlin-etal-2019-bert}, and proposed data augmentation with a contrastive loss for pretraining more robust multilingual models.
More recently, \citet{aliakbarzadeh2025exploringrobustnessmultilingualllms} extended this line of investigation to larger multilingual models, demonstrating that performance deteriorates when inputs are corrupted by real-world noise in language understanding tasks.
In contrast to them, we explore typographical errors directly by introducing a multilingual, keyboard-aware typo generation algorithm that enables \emph{realistic} and \emph{extensible} simulation across diverse languages.
Additionally, our systematic evaluation covers a broader range of tasks beyond language understanding.


\section{\algorithmname}\seclabel{multypo}

This section introduces \algorithmname, a multilingual typo generation algorithm designed to simulate realistic, human-like typos based on language-specific keyboard layouts.
Given a clean input text, \algorithmname injects character-level perturbations that mimic natural typing mistakes, producing corrupted outputs that maintain the overall coherence of the original text.
We describe the algorithmic design in \secref{algorithm_design}, followed by a human evaluation in \secref{user_study} that evaluates how well the generated typos reflect real human typing behavior.

\subsection{Algorithm Design}\seclabel{algorithm_design}

To reflect the real typing behavior of users of different languages, we leverage a keyboard layout database.\footnote{\url{https://kbdlayout.info/}}
When inserting typos, special symbols (e.g., punctuation marks or modifier keys such as \emph{Enter}) are excluded.
For determining which hand types a specific key, we rely on the standard \emph{10-finger typing convention} for QWERTY English keyboards, according to which characters such as ``5TGB'' are assigned to the left hand and
``6YHN'' to the right hand \citep{logan2016different}. 
For other languages, we adopt the same keyboard-relative hand separation (i.e., left vs. right half, based on key positions) -- a common implicit assumption in multilingual layout design -- though we validate its appropriateness empirically in our human evaluation.
The overall pipeline of \algorithmname is illustrated in Figure~\ref{fig:pipeline}. 
Below, we describe the key components in detail, and the description of \algorithmname.

\paragraph{Typo Types}
We consider four types of typos based on existing literature, as introduced in \secref{background_typos}: \emph{replacement}, \emph{insertion}, \emph{deletion}, and \emph{transposition}.
\begin{itemize}

\item \textbf{\emph{Replacement}}: A single character in a word is replaced by a neighboring key based on the language-specific keyboard layout.

\item \textbf{\emph{Insertion}}: A randomly selected additional character is inserted immediately after a correctly typed character, simulating accidental simultaneous keystrokes.

\item \textbf{\emph{Deletion}}: A single character in a word is randomly deleted from the word, simulating the common case where a keypress is missed.

\item \textbf{\emph{Transposition}}: Two adjacent characters are swapped. 
We constrain this to occur only between characters typed with \emph{different} hands, based on the 10-finger typing convention.

\end{itemize}

\paragraph{Ignoring String Sets} 

To avoid corrupting tokens that are critical for downstream understanding, especially numbers, we define a language-specific set of strings to ignore during typo insertion.\footnote{While this makes the noise slightly less realistic, it ensures that benchmark results reflect robustness to typos rather than being skewed by altered numeric values in the prompt.}
These sets include numerical expressions commonly used across languages -- both in digit form (e.g., 1, 2, 3) and in word form (e.g., ``three'', ``hundred'', ``million'') (see Figure~\ref{fig:ignoring_set} in \secref{algorithm_detail}).
During typo generation, any word that matches or contains a string in the set is excluded from being inserted with typos.

\paragraph{Length-Aware Sampling Probability}

Rather than treating all words equally, we assign each word a sampling probability proportional to the square root of its length: $\frac{\sqrt{|w|}}{\sum_w{\sqrt{|w|}}}$ (normalized over all words in a given text), reflecting the tendency for longer words to attract more typos \citep{peterson1986note,kukich1992techniques}.
In addition, when selecting a specific character position within a word to insert or modify, we also consider position-dependent weights.
Following observations from \citet{lisbach2013linguistic}, which show that errors are more likely to occur toward the middle or end of a word, we assign a non-uniform probability distribution over character indices, with details provided in \secref{algorithm_detail}.

\paragraph{Algorithm Description} 
Given an input sequence $S = \{w_1, w_2, \ldots, w_n\}$ of $n$ words, our algorithm begins by computing the number of typos to insert, determined by a user-defined corruption ratio $\tau \in [0, 1]$ and the total number of words $n$, rounded to the nearest integer.
Each word $w_i$ is assigned a sampling probability proportional to the square root of its character length $\sqrt{|w|}$, as described earlier, and candidates are sampled accordingly.
For each selected word, we sample one of four typo operations: \emph{replace}, \emph{insert}, \emph{delete}, or \emph{transpose}.
The specific character position within the word is sampled based on a length-aware, position-dependent distribution that favors later positions.
Once the position in a word is determined, the algorithm applies the selected typo operation to that position.
After each successful typo insertion, the corresponding word’s sampling weight is halved to encourage distributional diversity.
The algorithm proceeds iteratively until either the target number of typos is reached or a maximum retry threshold is exceeded.
The pipeline of \algorithmname is illustrated in Figure~\ref{fig:pipeline}.

\begin{table}[t]
\centering
\footnotesize
\setlength{\belowcaptionskip}{-0.5cm}
\resizebox{\linewidth}{!}{
\begin{tabular}{lrrr}
\hline
\textbf{Language} & \textbf{Multypo (avg.)} & \textbf{Naive (avg.)} & \textbf{Significance} \\
\hline
Arabic     & 6.00  & 6.60  &     \\
Armenian   & 10.20 & 4.87  & *** \\
Bengali    & 8.87  & 6.20  & *** \\
English    & 10.00 & 4.79  & *** \\
French     & 9.60  & 6.67  & *** \\
Georgian   & 8.87  & 7.27  & *** \\
German     & 8.93  & 5.27  & *** \\
Greek      & 8.07  & 5.93  & *  \\
Hebrew     & 9.93  & 7.60  & *** \\
Hindi      & 9.40  & 6.67  & ** \\
Russian    & 9.67  & 7.67  & ** \\
Tamil      & 6.13  & 5.07  & *   \\
\hline
\end{tabular}
}
\caption{Average number of sentences judged as ``natural'' out of 15 corrupted sentences per system, with significance from paired t-tests. Stars denote the significance levels: * $p<0.05$, ** $p<0.01$, *** $p<0.001$.}
\label{tab:user_study_nat}
\end{table}

\subsection{Human Evaluation on Typo Naturalness}\seclabel{user_study}

To further assess the realism of \algorithmname-generated typos, we conduct a human evaluation comparing \algorithmname against a \emph{naive baseline} that applies the same four operations (\emph{insertion}, \emph{deletion}, \emph{substitution}, \emph{transposition}) but without considering keyboard layout constraints. 
For each language, we sample 30 sentences from Flores200 \citep{flores200}, split into two equal halves: 15 sentences corrupted by \algorithmname and 15 by the naive baseline. 
Within each set, we balanced the number of sentences across 3 corruption levels ($0.1$, $0.4$, and $0.7$; five sentences per level), while ensuring that sentence length distributions remained comparable across the two conditions.
At least 15 participants in each language were asked to judge whether the typos in each sentence appeared \emph{natural} or \emph{unnatural}. 
This binary judgment provides a direct measure of how human-like the errors appear. 
We collected annotations across seven languages: Arabic, German, Greek, English, French, Hindi, and Russian (details provided in \secref{human_eva}).

Table~\ref{tab:user_study_nat} summarizes the results across languages. 
In six of the seven cases, \algorithmname is judged significantly more natural than the random baseline in the paired t-test (at least $p<0.05$). 
Arabic is the only exception, where ratings slightly favored the baseline, but without statistical significance.
We include it in our further experiments for completeness, though results might need to be interpreted with caution.
Taken together, this human evaluation confirms that \algorithmname can generally generate typos perceived as more human-like across languages than a naive baseline process that does not consider keyboard layout constraints.
In \secref{comparison_baseline}, we also show that models exhibit different robustness to our typos versus the naive ones: models are more robust to typos generated by \algorithmname, possibly due to the exposure of similar typos -- real-world human typos -- in the pretraining phase.

\section{Experimental Setup}


This section outlines our evaluation setup, where we apply \algorithmname to inject human-like typos into diverse downstream tasks and assess the robustness of different LLMs to these perturbations.

\subsection{Languages}

We consider 12 languages spanning 7 language families and written in 7 different scripts, with a focus on alphabet-based writing systems where typos are primarily influenced by keyboard layout.
The set of supported languages by \algorithmname includes Arabic (\textbf{ara\_Arab}), Armenian (\textbf{hye\_Armn}), Bengali (\textbf{ben\_Beng}), English (\textbf{eng\_Latn}), French (\textbf{fra\_Latn}), Georgian (\textbf{kat\_Geor}), German (\textbf{deu\_Latn}), Greek (\textbf{ell\_Grek}), Hebrew (\textbf{heb\_Hebr}), Hindi (\textbf{hin\_Deva}), Russian (\textbf{rus\_Cyrl}), and Tamil (\textbf{tam\_Taml}).

\subsection{Models} 
We evaluate 18 decoder-only language models from 3 model families:
\textbf{Gemma} \citep{gemmateam2025gemma3technicalreport},
\textbf{Qwen} \citep{yang2025qwen3technicalreport}, and \textbf{OLMo} \citep{olmo20252olmo2furious}. 
Models from the first two families are pretrained on highly multilingual corpora, while OLMo is pretrained on English-centric data. 
For the Gemma family, we consider \texttt{gemma-3-1b-pt}, \texttt{gemma-3-4b-pt}, and \texttt{gemma-3-12b-pt}.
For the Qwen family, we consider \texttt{Qwen3-1.7B-Base}, \texttt{Qwen3-4B-Base}, and \texttt{Qwen3-8B-Base}. 
Finally for the OLMo family, we consider \texttt{OLMo-2-0425-1B}, \texttt{OLMo-2-1124-7B}, and \texttt{OLMo-2-1124-13B}.
For each model above, we also consider its corresponding instruction-tuned version, aiming to systematically investigate the robustness against multilingual typos of models across size, family, and training strategies.

\subsection{Dataset}

\begin{figure*}
    \centering
    \includegraphics[width=0.31\textwidth]{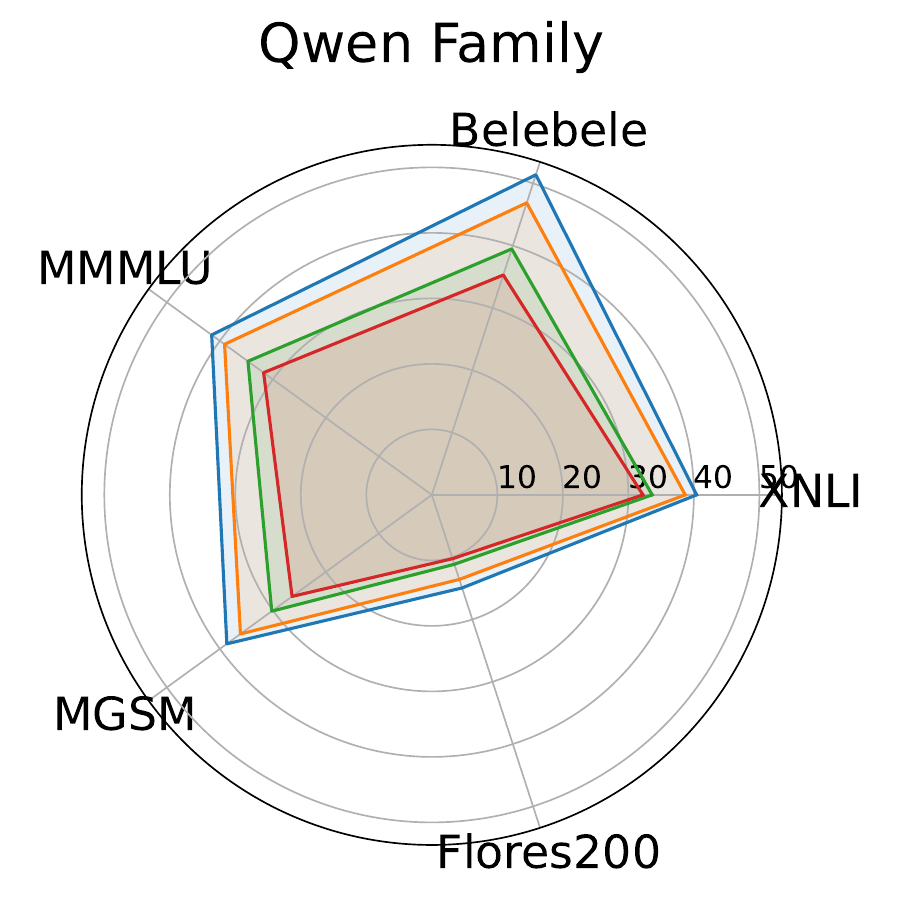}
    \includegraphics[width=0.31\textwidth]{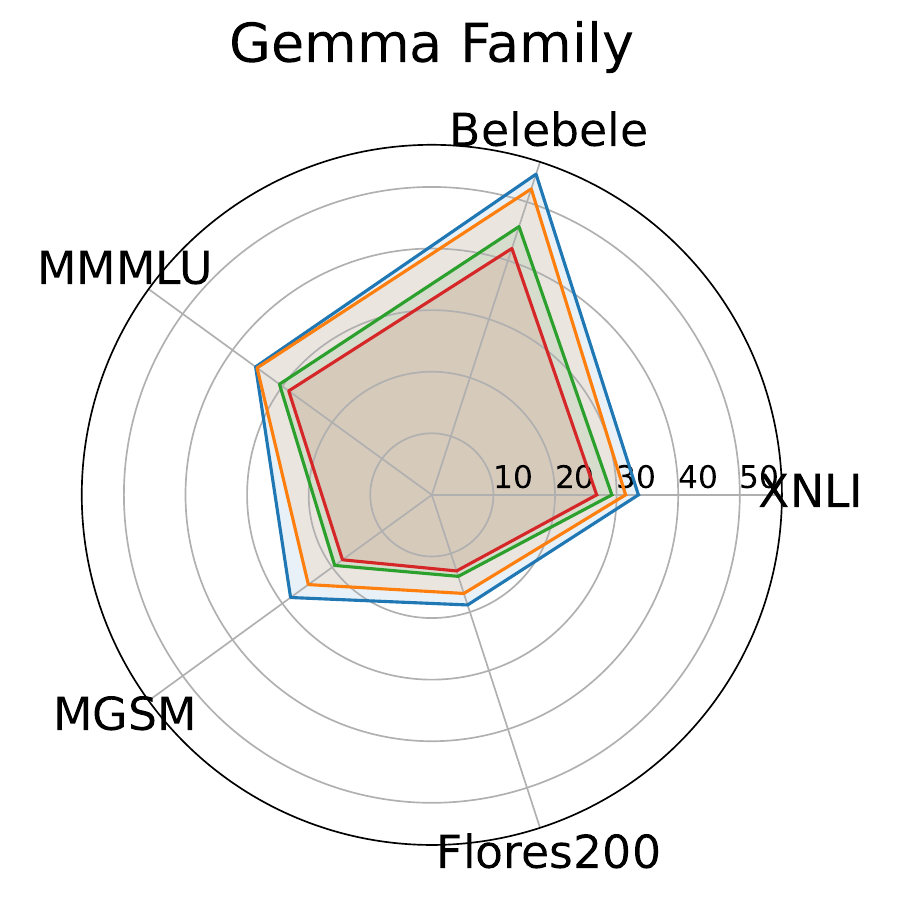}
    \includegraphics[width=0.31\textwidth]{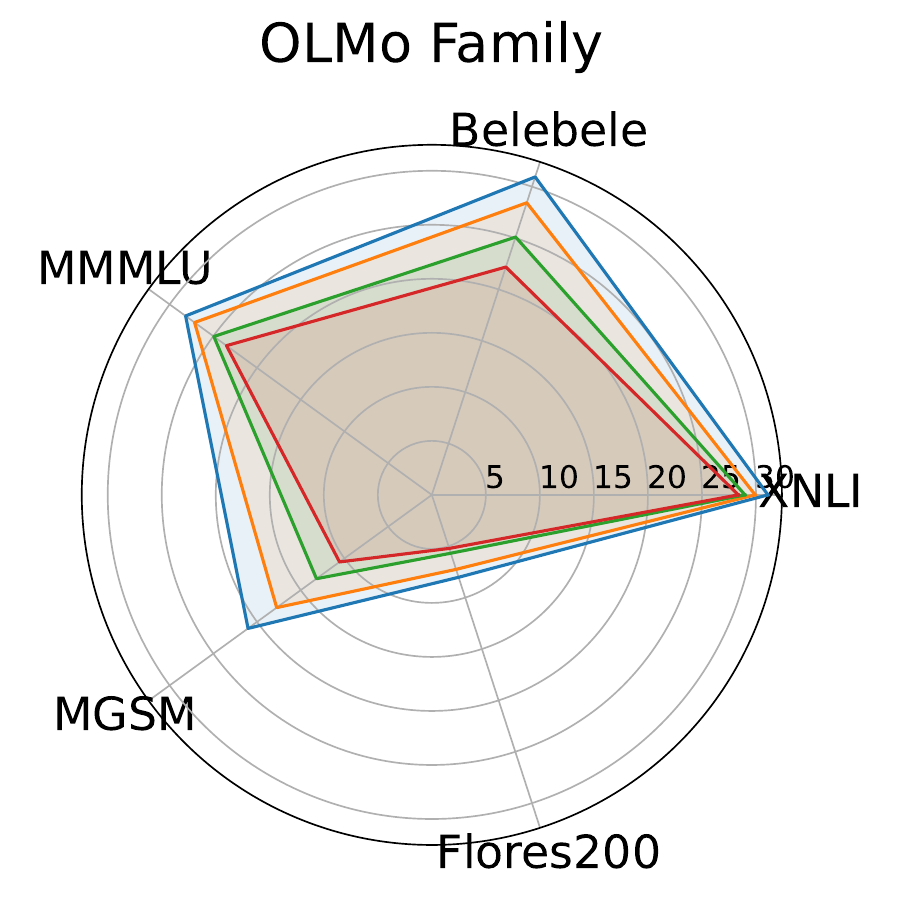}
    \includegraphics[width=0.60\textwidth]{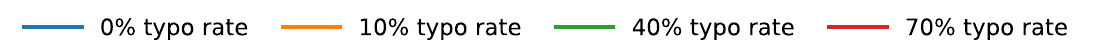}
    \caption{Performance under different typo rates ($0$, $0.1$, $0.4$, and $0.7$), averaged across languages for each task and model family. 
    Performance consistently degrades as the typo rate increases across all task types. 
    Notably, tasks requiring reasoning (e.g., MGSM) exhibit larger performance drops, indicating higher vulnerability to input noise.}
    \label{fig:typo_rate}
\end{figure*}

\begin{figure*}
    \centering
    \setlength{\belowcaptionskip}{-0.5cm}
    \includegraphics[width=0.19\textwidth]{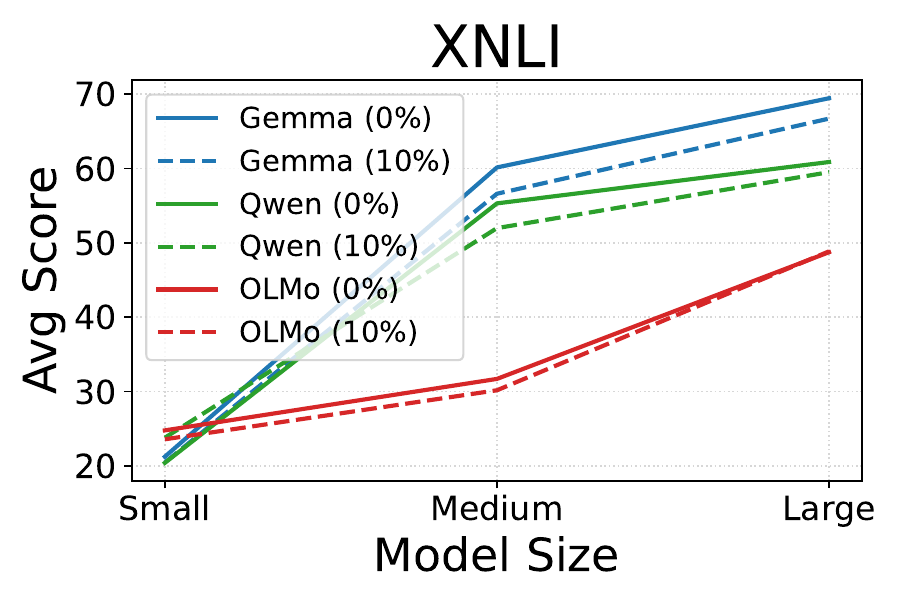}
    \includegraphics[width=0.19\textwidth]{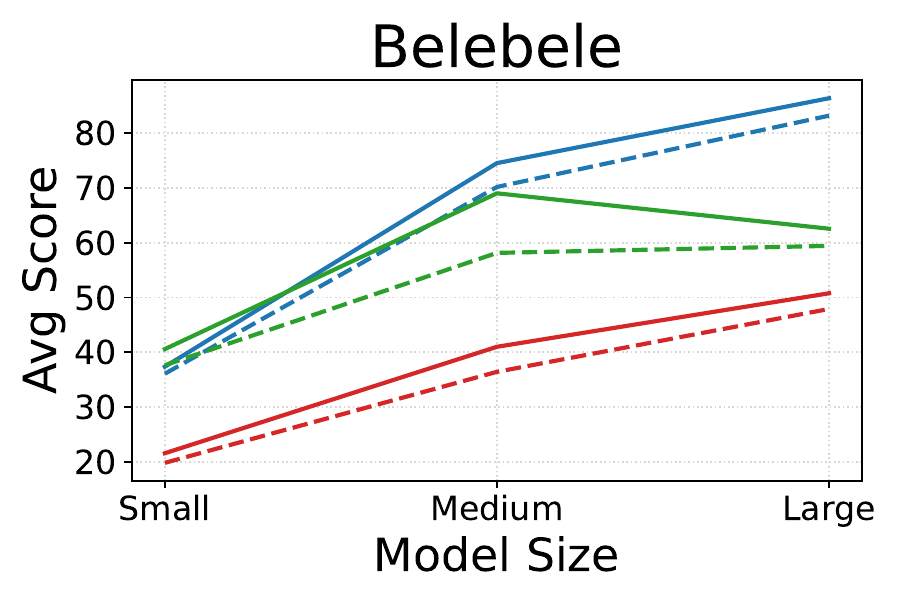}
    \includegraphics[width=0.19\textwidth]{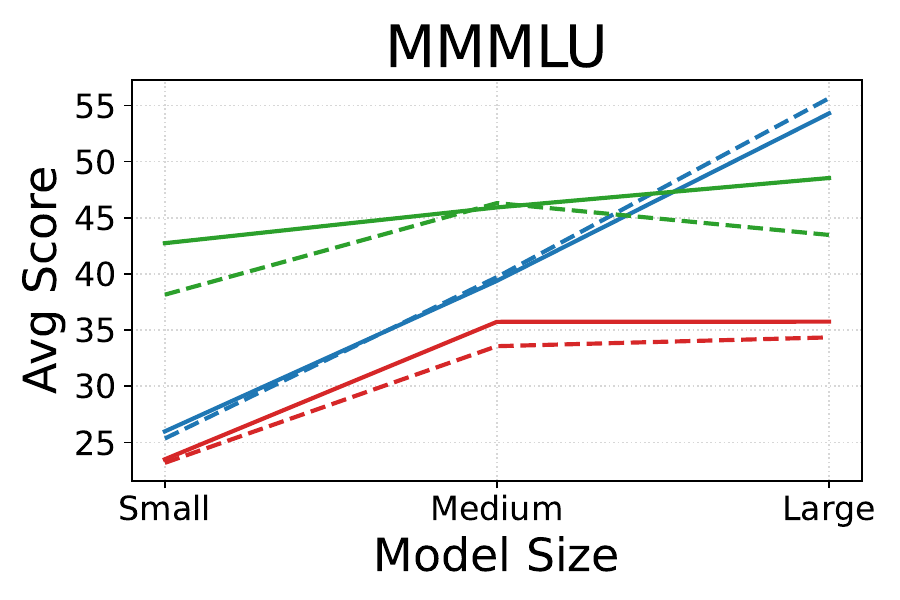}
    \includegraphics[width=0.19\textwidth]{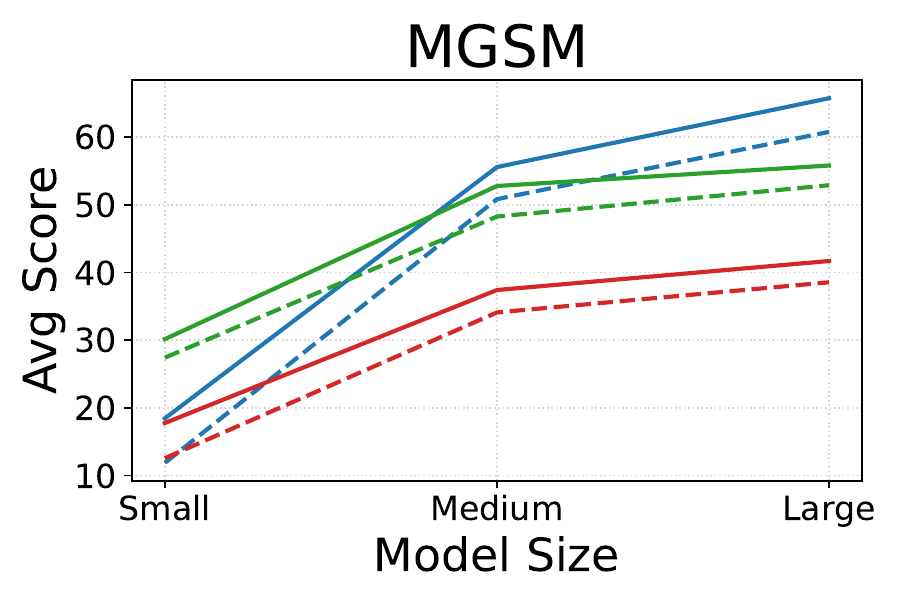}
    \includegraphics[width=0.19\textwidth]{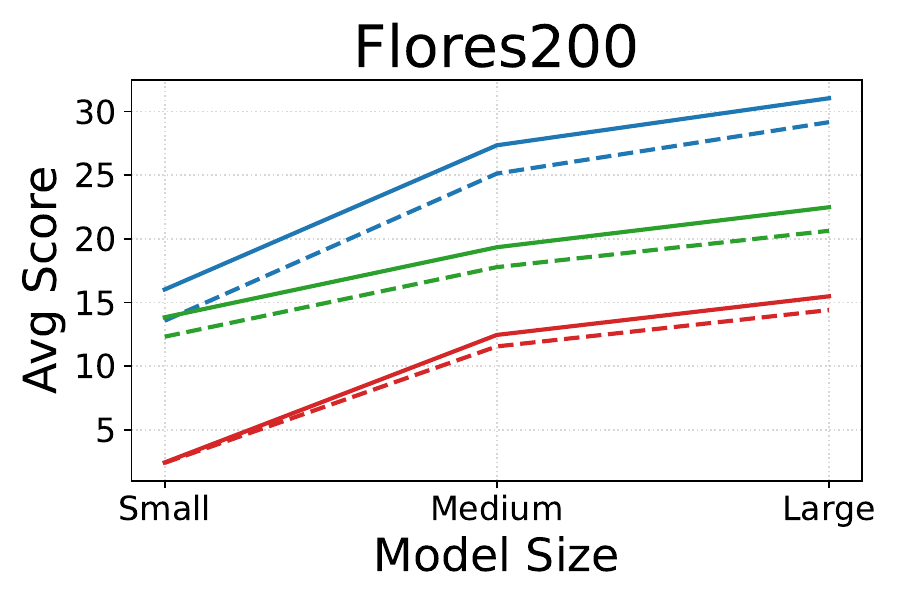}
    \caption{Impact of model size (Small, Medium, Large) on multilingual robustness across five tasks. 
    A different color represents each model family, and two lines are plotted per family: performance on clean input (0\%) and input with a 10\% typo rate. 
    Larger models generally perform better but also exhibit performance drops under noise.}
    \label{fig:model_size}
\end{figure*}

To evaluate robustness under multilingual typos, we use six datasets spanning four task types: natural language inference (\textbf{XNLI} \citep{conneau-etal-2018-xnli}), multiple-choice questions answering (\textbf{Belebele} \citep{bandarkar-etal-2024-belebele} and \textbf{MMMLU} \citep{mmmlu}), mathematical reasoning (\textbf{MGSM} \citep{mgsm}, along with Arabic and Hindi adaptations of GSM8K \citep{GSM8K,hindi-gsm8k,arabic-gsm8k}), and machine translation (\textbf{FLORES200} \citep{flores200}). 
These datasets are selected to cover diverse tasks and our target languages.
Note that the typos are only injected into the dataset instances, but not into other components of the prompt, such as task instructions, to ensure that we are evaluating robustness to input corruption rather than altering the task specification itself.
Further details, including language coverage of each dataset and used prompt templates, are provided in \secref{downstream_detail}.

\section{Results and Discussion}

In this section, we present the results of injecting typos into different datasets.
By default, we use 3-shot prompting to ensure a reasonable performance (we further analyze the effect of example-count on robustness in \secref{shot_count}).
In the following parts, we aim to investigate three research questions: \textbf{(1)} \emph{Does performance degrade when typographical errors are introduced, and if so, how much?} (\secref{typo_rate}); \textbf{(2)} \emph{Do larger models present better robustness against typographical errors compared to smaller ones?} (\secref{model_size}); and \textbf{(3)} \emph{Does instruction-tuning improve the robustness of the models?} (\secref{instruction_tuned}).

\subsection{Performance Drop under Typos}\seclabel{typo_rate}

Figure~\ref{fig:typo_rate} presents the average performance of three model families -- {Gemma}, {Qwen}, and {OLMo} -- across five multilingual tasks, under varying levels of typographical corruption (0\%, 10\%, 40\%, 70\%). 
Results are aggregated over instruction-tuned models and all supported languages within each task.

\textbf{Typos consistently degrade performance across all models.}
Across all families and tasks, even minor typographical noise largely impairs model performance. 
For example, Qwen achieves over 50 on {Belebele} in the clean setting, but drops to around 45 with just a 10\% typo rate.
As noise increases, performance declines continuously. 
This pattern holds across families and tasks, underscoring a general vulnerability to surface-level 
perturbations. 
These findings echo prior monolingual results \citep{moradi-samwald-2021-evaluating,wang2025perturbations}, and extend them to a multilingual setting.

\textbf{Robustness varies substantially by task.}
Typo sensitivity is not uniform across tasks. 
For instance, XNLI exhibits good robustness: Qwen's performance remains nearly unchanged under 10\% noise. 
Even the OLMo models -- despite being primarily monolingual -- sustain less than a 10-point absolute drop at the highest noise level. 
In contrast, tasks involving generative reasoning (e.g., MGSM) are highly susceptible.
Qwen’s accuracy on MGSM plummets from around 40 (clean) to around 27 (70\% noise), suggesting that token-level corruption disrupts multi-step reasoning more than classification-based understanding. 
These results support earlier, monolingual claims that noisy inputs affect complex reasoning \citep{gan-etal-2024-reasoning}.

\paragraph{Takeaway.}
While LLMs can often infer intended meaning from noisy input in simpler classification tasks (e.g., natural language inference), reasoning tasks amplify the fragility introduced by typos. 

\subsection{Model Size Impact on Robustness}\seclabel{model_size}

\begin{table}
    \centering
    \setlength{\belowcaptionskip}{-0.4cm}
    \footnotesize
    \resizebox{\linewidth}{!}{%
    \begin{tabular}{lrrr}
    \toprule
    Family & Small & Medium & Large \\
    \midrule
    Gemma & 21.46 (-9.9\%) & 48.50 (-5.7\%) & 59.11 (-3.7\%) \\
    OLMo & 16.30 (-9.5\%) & 29.16 (-7.9\%) & 36.82 (-4.3\%) \\
    Qwen & 27.86 (-5.7\%) & 44.50 (-8.2\%) & 47.19 (-5.7\%) \\
    \bottomrule
    \end{tabular}
    }
    \caption{
    Each cell reports the average score of a model family under a 10\% typo rate, with the relative performance drop from clean input shown in parentheses.}
    \label{tab:model_size}
\end{table}

Figure~\ref{fig:model_size} presents the average performance when fed with clean inputs and with a small typo rate (10\%). 
We group the models in each model family into different scales (Small, Medium, and Large).
Results are averaged across tasks and supported languages, focusing on instruction-tuned models.

\begin{figure*}
    \centering
    \setlength{\abovecaptionskip}{-0.05cm}
    \setlength{\belowcaptionskip}{-0.5cm}
    \includegraphics[width=0.31\textwidth]{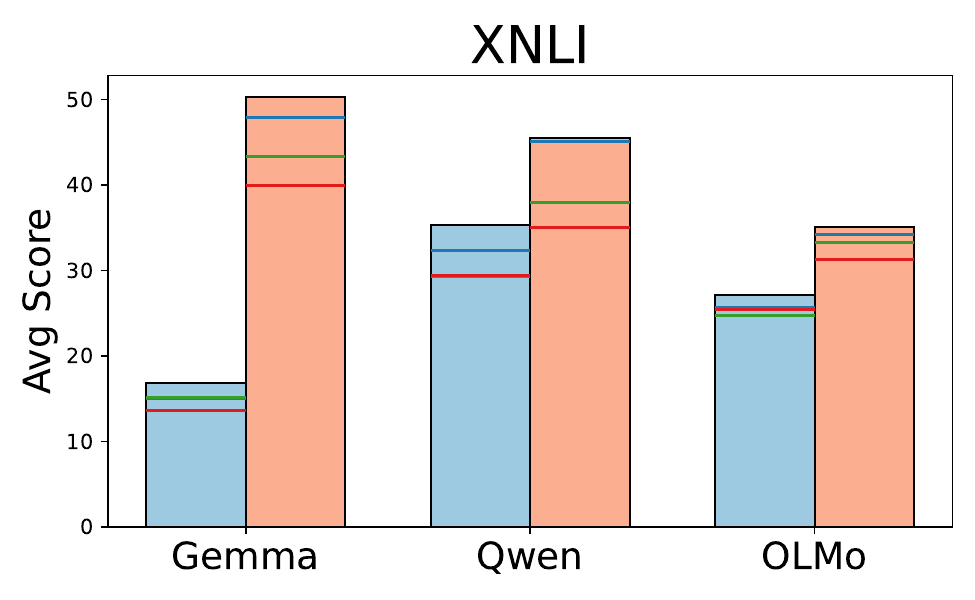}
    \includegraphics[width=0.31\textwidth]{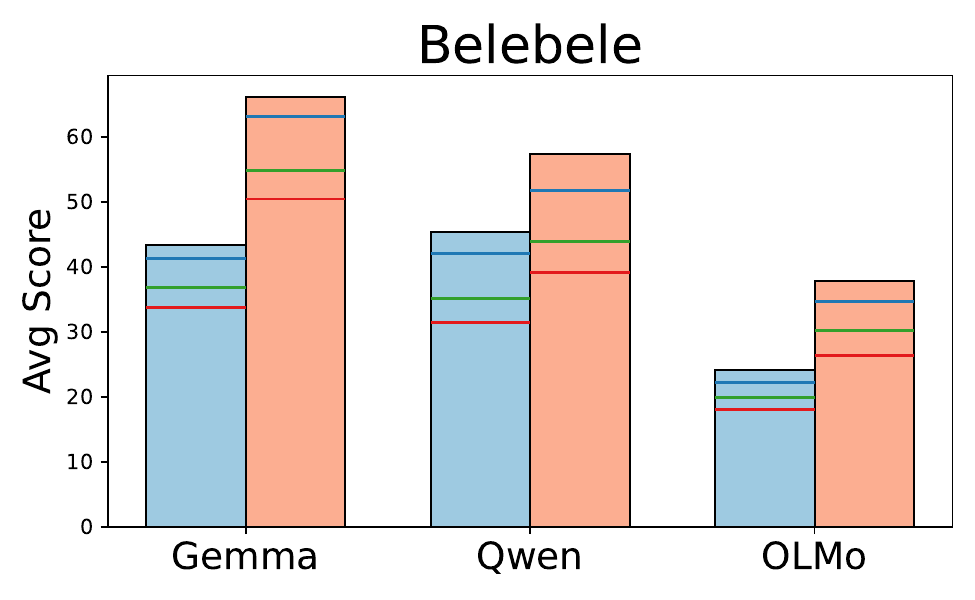}
    \includegraphics[width=0.31\textwidth]{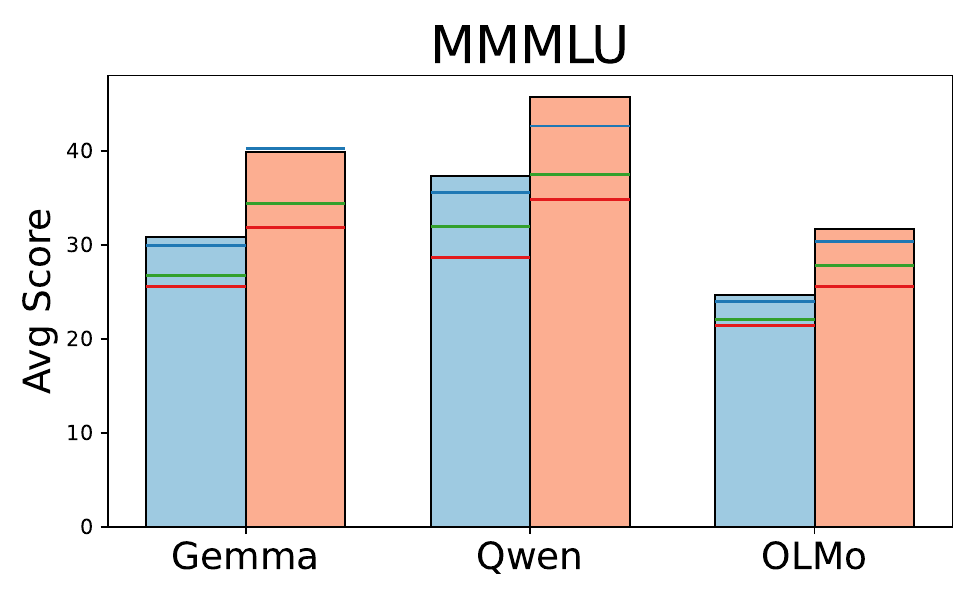}
    \includegraphics[width=0.31\textwidth]{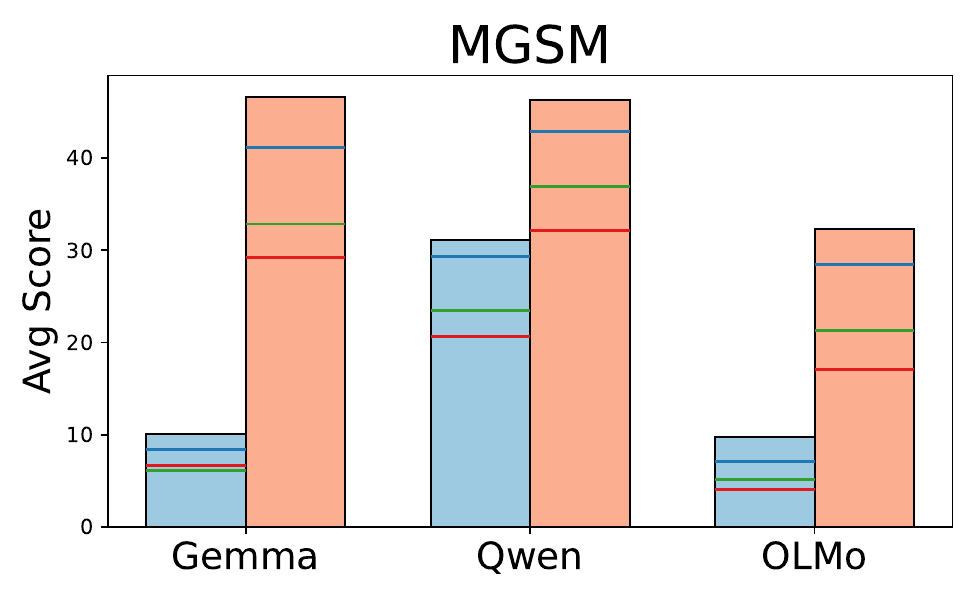}
    \includegraphics[width=0.31\textwidth]{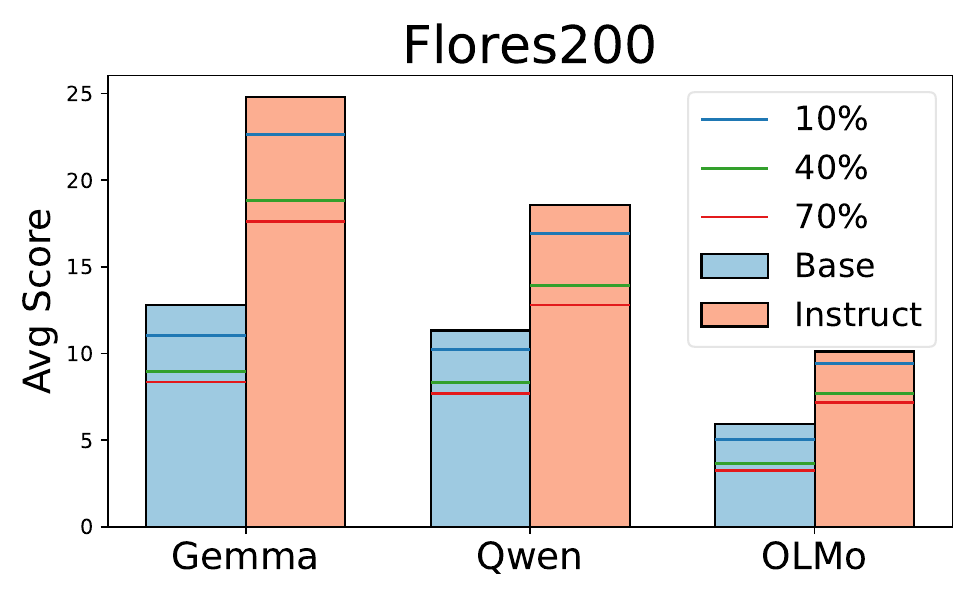}
    \caption{Impact of Instruction-tuning on multilingual robustness.
    Instruction-tuned models improve the performance, but do not seem to improve the robustness against typos, especially with higher typo rates.}
    \label{fig:instruct_vs_base}
\end{figure*}

\textbf{Larger models consistently outperform smaller ones, with mild gains in robustness.}
While larger models consistently outperform smaller ones, also under typo noise, models of all sizes suffer from input perturbations, as shown in Figure~\ref{fig:model_size}.
To further analyze the effect of model scale on robustness, we compute the relative degradation under a 10\% typo rate ($\frac{\text{Perf}_{10\%} - \text{Perf}_{0\%}}{\text{Perf}_{0\%}}$). 
As reported in Table~\ref{tab:model_size}, larger models show smaller drops -- particularly within the Gemma and OLMo families. 
E.g., Gemma's relative drop decreases from 9.9\% (Small) to 3.7\% (Large), suggesting that greater model capacity enables better robustness. 


\paragraph{Takeaways.}
Scaling the model improves task performance under typo noise, but larger models are not immune to noise as well.
However, larger models present improved robustness under typos.

\subsection{Base vs. Instruction-Tuned Models}\seclabel{instruction_tuned}

Figure~\ref{fig:instruct_vs_base} presents the performance of pretrained based models and their instruction-tuned versions under varying levels of typographical corruption (0\%, 10\%, 40\%, 70\%).
Results are averaged over all supported languages within each task.


\textbf{Instruction-tuned models outperform base models but remain brittle under typos.}
Instruction-tuning improves overall performance, aligning with prior work \citep{liu-etal-2023-logicot,Won2o24scaling}, which shows that instruction-tuning enhances task-specific multi-step reasoning.
Despite clear performance benefits under clean input, instruction-tuned models remain vulnerable to typos.
In many cases, the absolute degradation under 10\% or 40\% noise is as severe as or even worse than their base counterparts.
For instance, on \textsc{MGSM}, Gemma's instruction-tuned models drop from around 48 to 33 under 40\% corruption.
Similar degradation is seen across other families and tasks. 
This suggests that while instruction-tuned models are better at following complex prompts, they remain equally brittle under surface-level input corruption.

\paragraph{Takeaways.}
Instruction-tuning boosts performance but does not improve robustness. 
Current tuning methods prioritize clean prompts and may underprepare models for noisy real-world input.


\section{Complementary Analysis}

\subsection{Comparison with \textsc{WikiTypos}}

\begin{table}[t]
\centering
\footnotesize
\setlength{\belowcaptionskip}{-0.5cm}
\setlength{\tabcolsep}{4pt}
\resizebox{\linewidth}{!}{
\begin{tabular}{lccccc}
\toprule
\textbf{Method (typo rate)} & \textbf{XNLI} & \textbf{Belebele} & \textbf{MMMLU} & \textbf{MGSM} & \textbf{Flores200} \\
\hline
Baseline (10\%)   & 56.25 & 74.83 & 35.73 & 46.90 & 35.47 \\
\textsc{WikiTypo} (10\%) & 57.65 & 73.07 & 37.80 & 53.30 & 35.20 \\
\algorithmname (10\%)  & 55.83 & 76.58 & 43.43 & 53.80 & 35.35 \\
\hline
Baseline (40\%)   & 48.83 & 63.20 & 33.20 & 26.00 & 30.16 \\
\textsc{WikiTypo} (40\%) & 50.20 & 63.62 & 33.90 & 36.20 & 28.12 \\
\algorithmname (40\%)  & 49.30 & 64.90 & 37.52 & 42.70 & 31.39 \\
\hline
Baseline (70\%)   & 40.67 & 56.20 & 30.62 & 12.00 & 24.21 \\
\textsc{WikiTypo} (70\%) & 38.80 & 52.65 & 30.45 & 16.40 & 22.47 \\
\algorithmname (70\%)  & 43.20 & 61.85 & 31.27 & 38.80 & 29.68 \\
\hline
\end{tabular}
}
\caption{Performance comparison under different typo generation methods and typo rates. 
Performance consistently degrades as the typo rate increases across all methods. 
\algorithmname generally exhibits degradation patterns that lie between Baseline and \textsc{WikiTypos}.}
\label{tab:wikitypo}
\end{table}

\begin{figure*}
    \centering
    \setlength{\abovecaptionskip}{-0.05cm}
    \setlength{\belowcaptionskip}{-0.3cm}
    \includegraphics[width=0.32\textwidth]{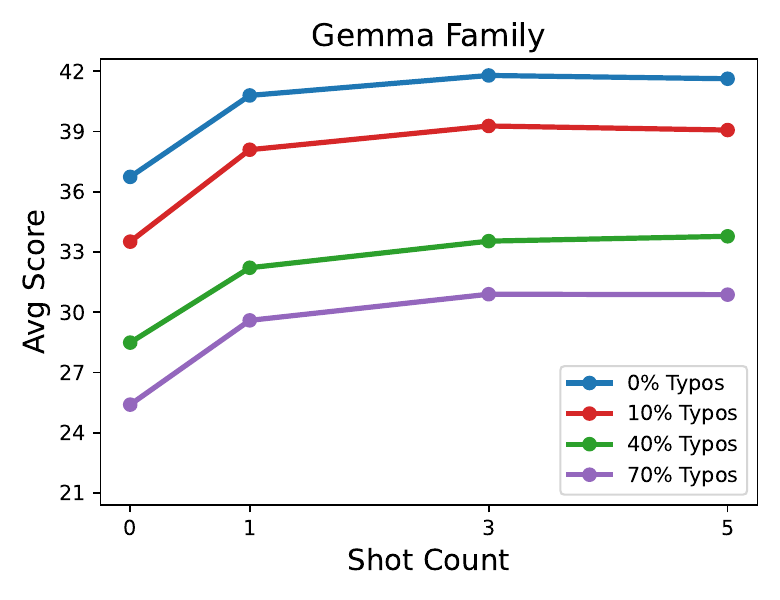}
    \includegraphics[width=0.32\textwidth]{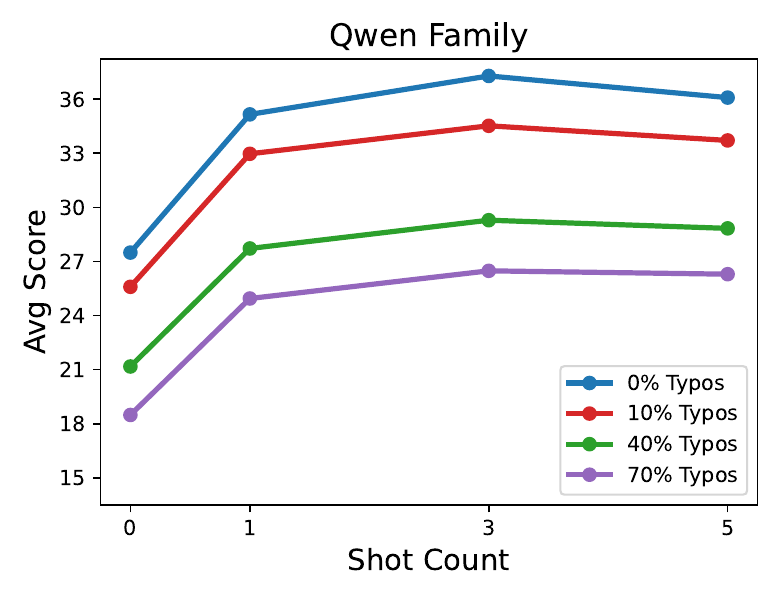}
    \includegraphics[width=0.32\textwidth]{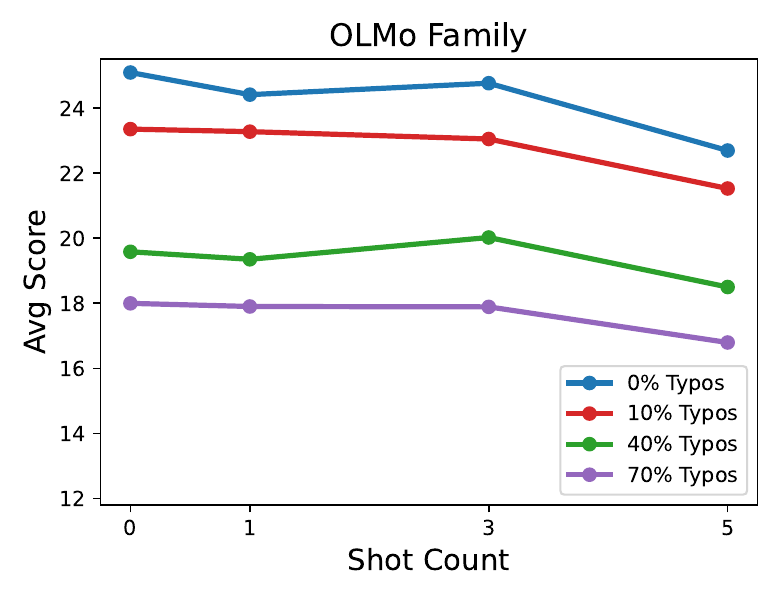}
    \caption{Performance of different model families under different numbers of shots.
    Increasing shot counts, i.e., the number of demonstrations, slightly improves the performance but does not improve the robustness against typos.}
    \label{fig:shot_count}
\end{figure*}

To further validate the realism of \algorithmname, we compare it with \textsc{WikiTypos} \citep{aliakbarzadeh2025exploringrobustnessmultilingualllms}, a multilingual dataset of real-world edits extracted from Wikipedia edit history.\footnote{\textsc{WikiTypos} also includes semantic substitutions (e.g., replacing a word with another of different meaning), introducing noise beyond typographical errors.}
We extract typo pairs from \textsc{WikiTypos} and inject them into our evaluation datasets. 
As an additional baseline, we include a naive typo generation method that applies the same four operations described in \secref{algorithm_design}, but without considering keyboard layout constraints. 
We conduct experiments on four overlapping languages (English, German, French, Hindi) using \texttt{gemma-3-4b-it}.
Evaluation is performed across five tasks (XNLI, Belebele, MMMLU, MGSM, Flores200) under varying typo rates (10\%, 40\%, 70\%).

\begin{figure}[t]
    \centering
    \setlength{\belowcaptionskip}{-0.3cm}
    \includegraphics[width=0.9\linewidth]{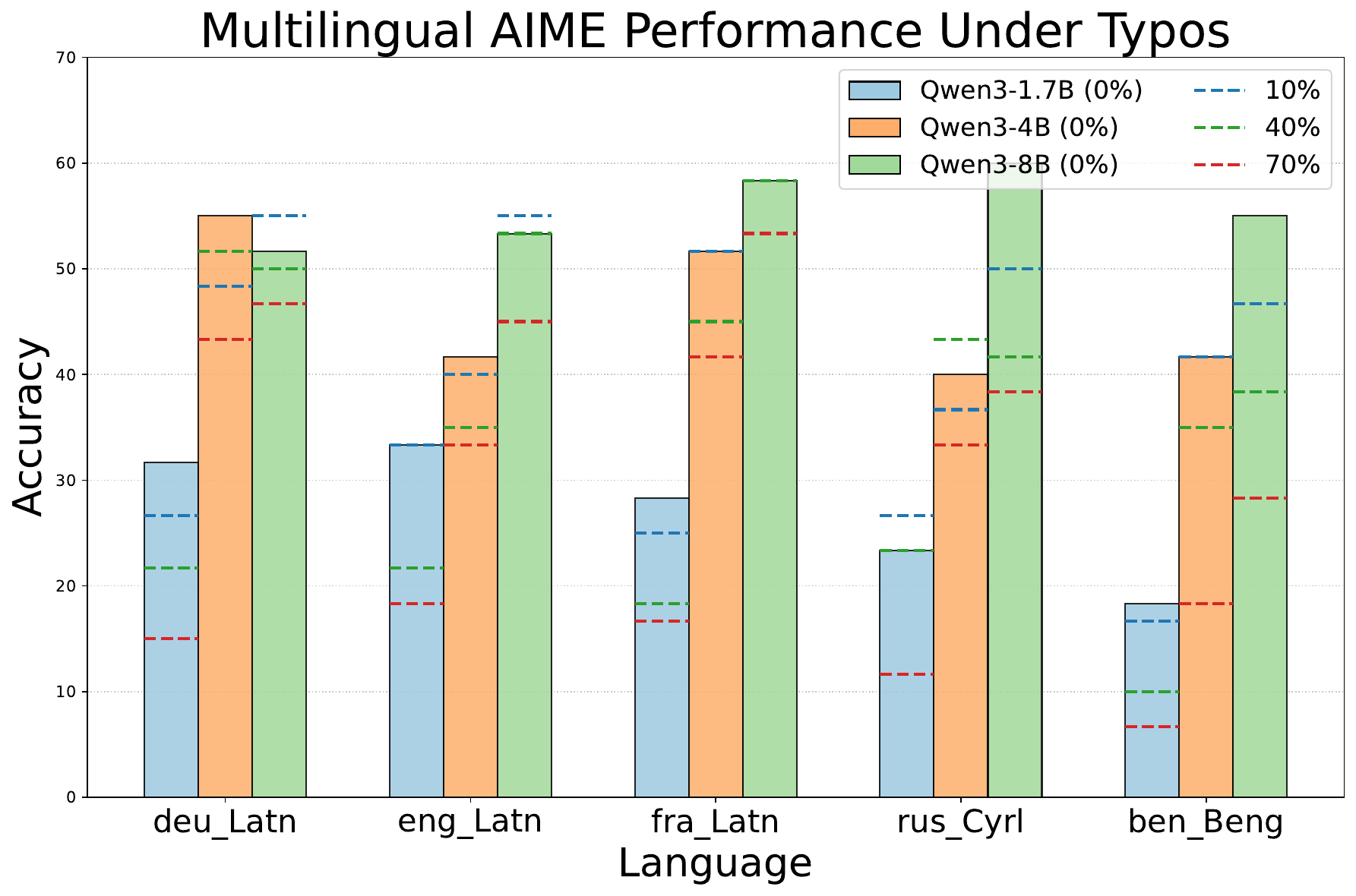}
    \caption{
    Performance on multilingual AIME, a harder mathematical reasoning benchmark, under different typo rates across five languages. Accuracy generally decreases as the typo rate increases, while larger models remain more robust than smaller ones. Crosslingual differences are also visible, with Bengali and Russian (which use scripts other than the Latin script) typically exhibiting greater degradation under heavy noise.}
    \label{fig:aime_typo_barplot}
\end{figure}

Table~\ref{tab:wikitypo} presents the results across methods and typo rates. 
We observe that performance consistently degrades as the typo rate increases for all methods, indicating that the degree of textual perturbation strongly affects model performance. 
Furthermore, \algorithmname generally exhibits degradation patterns that lie between the naive baseline and \textsc{WikiTypos} across many tasks. 
This suggests that models are relatively more robust to \algorithmname and \textsc{WikiTypo} perturbations, likely because these better reflect realistic human typing errors, some of which may already be present in pretraining data. 
\textbf{Overall, these findings further support the validity of using \algorithmname as a realistic and controlled approach for evaluating model robustness under real-world noisy inputs.}

\subsection{Influence on Harder Benchmarks}

To examine whether our findings generalize to more challenging reasoning tasks, we extend our evaluation to multilingual AIME, constructed by combining AIME 2025 and AIME 2026 \citep{qi-etal-2025-models}, yielding a total of 60 math problems per language. 
Compared to MGSM, AIME requires substantially longer and more complex reasoning traces and is less likely to suffer from data contamination. 
We evaluate the Qwen3 family (1.7B, 4B, 8B) across five languages (German, English, French, Russian, Bengali) under varying typo rates (0\%, 10\%, 40\%, 70\%).

Figure~\ref{fig:aime_typo_barplot} presents the results. 
Our key findings consistently generalize to this more challenging benchmark. 
As the typo rate increases, performance declines across nearly all models and languages, indicating that complex reasoning tasks also remain highly sensitive to input noise. 
While larger models (e.g., Qwen3-8B) exhibit stronger robustness than smaller ones, they still experience noticeable degradation under higher noise levels. 
We further observe clear cross-lingual variation: high-resource languages written in the Latin script (English, German, French) tend to be more stable, whereas Russian and especially Bengali show larger performance drops as the typo rate increases.
This pattern aligns with prior findings that chain-of-thought reasoning is more robust in high-resource languages \citep{zhao-etal-2026-comprehensive,liu2026largereasoningmodelsnot}.
\textbf{Our results further confirm that tasks involving complex reasoning are particularly susceptible to typos, and that this vulnerability may even be amplified in more challenging benchmarks.}

\begin{figure*}
    \centering
    \setlength{\abovecaptionskip}{-0.05cm}
    \setlength{\belowcaptionskip}{-0.3cm}
    \includegraphics[width=0.95\textwidth]{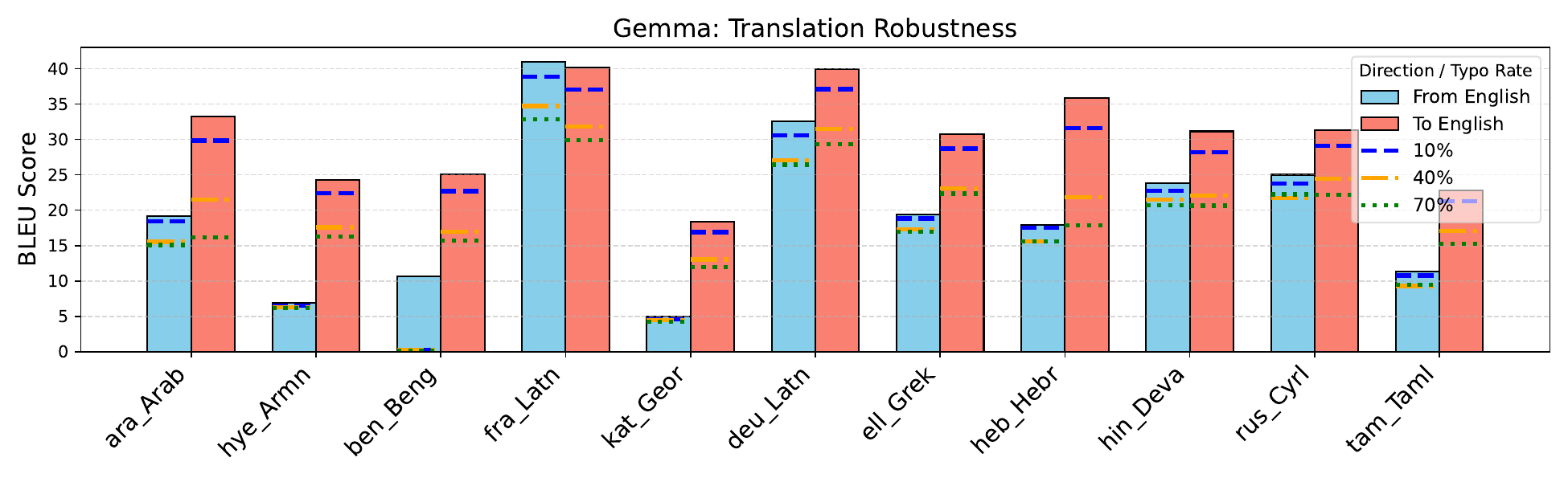}
    \caption{Robustness of \textbf{Gemma} models on \textbf{Flores200} under different levels of typographical noise.
    Translation from English seems to be more robust compared to translation to English.}
    \label{fig:flores_gemma}
\end{figure*}

\subsection{Do Example Counts Affect Robustness?}\seclabel{shot_count}

Few-shot prompting is known to enhance model performance by providing clearer task formulations and patterns \citep{Brown2020fewshot,schick-schutze-2022-true}. This naturally raises the question: \emph{Can increasing the number of examples also improve robustness against typos?} 
To explore this, we vary the number of examples in the prompt -- 0, 1, 3, and 5 -- and evaluate instruction-tuned models across different typo rates, tasks, and languages. 
Figure~\ref{fig:shot_count} presents the aggregated results.

Across the multilingual models (Gemma and Qwen), increasing the number of examples in few-shot settings leads to consistent performance gains until 3 shots.
However, the robustness gap, i.e., the performance drop from clean to noisy inputs, remains nearly unchanged regardless of shot count. 
For the OLMo family, adding more examples does not seem to help and occasionally harms performance, likely due to its limited multilingual coverage and confusion introduced by prompts in languages that OLMo does not support well.
\textbf{These findings suggest that while demonstrations improve overall performance, they do not inherently enhance robustness against typos.}

\subsection{Which Language is More Sensitive?}\seclabel{language_sensitive}

\begin{table}[t]
\centering
\setlength{\belowcaptionskip}{-0.3cm}
\small
\renewcommand{\arraystretch}{1.2}
\setlength{\tabcolsep}{2pt}
\resizebox{\linewidth}{!}{
\begin{tabular}{lrrrrrrrr}
\toprule
\textbf{Family} & \textbf{ara\_Arab} & \textbf{ben\_Beng} & \textbf{deu\_Latn} & \textbf{ell\_Grek} & \textbf{eng\_Latn} & \textbf{fra\_Latn} & \textbf{hin\_Deva} & \textbf{rus\_Cyrl} \\
\midrule
\multirow{3}{*}{\textbf{Gemma}} & 51.3 & 47.3 & 53.0 & 55.3 & 56.4 & 53.6 & 45.6 & 57.3 \\
 & 46.3 & 42.3 & 52.2 & 55.4 & 57.5 & 52.1 & 41.3 & 50.5 \\
 & 9.7\% & 10.6\% & 1.5\% & -0.2\% & -2.0\% & 2.8\% & 9.4\% & 11.9\% \\
 \hline
\multirow{3}{*}{\textbf{Qwen}} & 47.2 & 38.3 & 52.9 & 47.6 & 62.3 & 54.8 & 41.0 & 55.0 \\
 & 44.6 & 31.6 & 50.8 & 44.5 & 58.8 & 53.0 & 36.9 & 52.4 \\
 & 5.5\% & 17.5\% & 4.0\% & 6.5\% & 5.6\% & 3.3\% & 10.0\% & 4.7\% \\
 \hline
\multirow{3}{*}{\textbf{OLMo}} & 30.6 & 20.3 & 40.9 & 39.1 & 57.5 & 46.3 & 26.9 & 41.9 \\
 & 28.5 & 19.1 & 38.3 & 37.1 & 55.8 & 43.4 & 24.6 & 35.3 \\
 & 6.9\% & 5.9\% & 6.4\% & 5.1\% & 3.0\% & 6.3\% & 8.6\% & 15.8\% \\
\bottomrule
\end{tabular}
}
\caption{Performance when fed with clean input (top row) and with a 10\% typo rate (bottom row), aggregated across all tasks and datasets, by language. Only languages supported by at least 3 datasets are considered.}
\label{tab:language-sensitivity}
\end{table}


We hypothesize that \emph{model robustness to typos is not uniform across languages}, particularly due to data availability. 
To investigate this, we analyze performance degradation across eight languages that are each supported by at least three datasets. 
Table~\ref{tab:language-sensitivity} presents results aggregated over all tasks. 
Additionally, we analyze \textbf{Flores200} separately to examine how translation direction interacts with typo robustness. 
Specifically, we compare the performance of Gemma models when translating \emph{from English} vs. \emph{into English}, as shown in Figure~\ref{fig:flores_gemma}.

Across all three model families, English consistently exhibits the highest robustness -- its relative drop is among the lowest. 
Other languages that use the Latin script, such as German and French, also show relatively small degradations. 
In contrast, languages with underrepresented scripts, including Arabic, Hindi, and Bengali, tend to exhibit larger drops. 
Interestingly, even Russian, despite being high-resource, suffers from sharp degradation (e.g., 11.9\% for Gemma).
This suggests that \textbf{models are more robust in languages with both high data availability and orthographic familiarity (e.g., Latin script)}. 
Furthermore, Figure~\ref{fig:flores_gemma} shows that translations \emph{from} English are more robust than those \emph{into} English, reinforcing the idea that \textbf{typos in lower-resourced or structurally different input languages more severely impair both crosslingual understanding and generation}.

\section{Conclusion}

We present a comprehensive study on the multilingual robustness of LLMs under simulated, realistic typographical errors. 
To this end, we introduce \algorithmname, a multilingual typo generation algorithm grounded in language-specific keyboard layouts and human typing behavior. 
Through extensive evaluation of 18 models across multiple tasks and three model families, we show that even modest levels of noise can substantially degrade performance -- particularly for reasoning-intensive tasks. 
While larger models and instruction tuning improve performance on clean inputs, they do not consistently translate to improved robustness under noise.
We further uncover crosslingual disparities: models are more resilient in high-resource, Latin-script languages, while exhibiting greater vulnerability in lower-resource or non-Latin-script languages.
These findings highlight critical blind spots in current LLM evaluation and motivate future work on noise-aware multilingual pretraining, evaluation, and human-centric error modeling.


\section*{Limitations}

While our work provides a first step toward multilingual robustness evaluation under human-like typographical errors, we acknowledge that several limitations remain.

First, \algorithmname currently supports a diverse but limited set of typologically diverse languages. 
To incorporate a new language, one needs to manually specify the corresponding keyboard layout and typing conventions.

Second, our algorithm does not yet support logographic or syllabic writing systems, such as Chinese.
This limitation stems from the fundamental differences in input methods -- e.g., Chinese characters are typically typed via phonetic systems like \emph{Pinyin} rather than direct keypresses. 
Modeling such input pipelines requires a fundamentally different corruption strategy. 
Future work could extend \algorithmname to accommodate these languages by simulating common typing errors in the intermediate input stages (e.g., Pinyin mistyping or candidate misselection).

Third, our human evaluation provides important validation of the realism of \algorithmname, covering multiple languages. 
However, results for Arabic did not show significant improvements over the naive baseline, suggesting that our typo simulation algorithm \algorithmname may not capture all language-specific properties equally well.
However, this does not overshadow the findings that LLMs are not robust to multilingual textual perturbations.

Finally, we focus exclusively on physical keyboards (e.g., QWERTY), while ignoring other input modalities such as touchscreen keyboards on mobile devices. Typing behaviors, error distributions, and auto-correct interference vary substantially across modalities \citep{jussi2021Touchscreen,shi2025simulating}.
Evaluating robustness under such device-dependent noise would further enrich our understanding of LLM performance in real-world settings, which we leave for future work.

\section*{Ethical Considerations}

\paragraph{Data Annotation} 

Before conducting the human evaluation, all participants were clearly informed about the purpose, procedure, and voluntary nature of the study, and provided their informed consent.
For most languages, annotators were recruited via Prolific and compensated fairly at a rate equivalent to approximately £6 per hour (about £1 per completed annotation set) (details are provided in \secref{human_eva}).
A small portion of participants (around 10\%) were personal contacts who volunteered without compensation.
No personally identifiable information was collected, and all demographic data (e.g., age, gender) was provided optionally.

\paragraph{Use of AI Assistants}
The authors acknowledge the use of ChatGPT exclusively for grammar correction, improving the clarity and coherence of the draft, and assisting with code implementation.\footnote{\url{https://chatgpt.com/}}

\section*{Acknowledgments}

This research was supported by the Munich Center for Machine Learning (MCML) and German Research Foundation (DFG, grant SCHU 2246/14-1).

\bibliography{custom}

@inproceedings{gan-etal-2024-reasoning,
    title = "Reasoning Robustness of {LLM}s to Adversarial Typographical Errors",
    author = "Gan, Esther  and
      Zhao, Yiran  and
      Cheng, Liying  and
      Yancan, Mao  and
      Goyal, Anirudh  and
      Kawaguchi, Kenji  and
      Kan, Min-Yen  and
      Shieh, Michael",
    editor = "Al-Onaizan, Yaser  and
      Bansal, Mohit  and
      Chen, Yun-Nung",
    booktitle = "Proceedings of the 2024 Conference on Empirical Methods in Natural Language Processing",
    month = nov,
    year = "2024",
    address = "Miami, Florida, USA",
    publisher = "Association for Computational Linguistics",
    url = "https://aclanthology.org/2024.emnlp-main.584/",
    doi = "10.18653/v1/2024.emnlp-main.584",
    pages = "10449--10459"
}

@misc{abedin2025arithmattackevaluatingrobustnessllms,
      title={ArithmAttack: Evaluating Robustness of LLMs to Noisy Context in Math Problem Solving}, 
      author={Zain Ul Abedin and Shahzeb Qamar and Lucie Flek and Akbar Karimi},
      year={2025},
      eprint={2501.08203},
      archivePrefix={arXiv},
      primaryClass={cs.CL},
      url={https://arxiv.org/abs/2501.08203}, 
}

@inproceedings{zhao-etal-2024-syntheval,
    title = "{S}ynth{E}val: Hybrid Behavioral Testing of {NLP} Models with Synthetic {C}heck{L}ists",
    author = {Zhao, Raoyuan  and
      K{\"o}ksal, Abdullatif  and
      Liu, Yihong  and
      Weissweiler, Leonie  and
      Korhonen, Anna  and
      Schuetze, Hinrich},
    editor = "Al-Onaizan, Yaser  and
      Bansal, Mohit  and
      Chen, Yun-Nung",
    booktitle = "Findings of the Association for Computational Linguistics: EMNLP 2024",
    month = nov,
    year = "2024",
    address = "Miami, Florida, USA",
    publisher = "Association for Computational Linguistics",
    url = "https://aclanthology.org/2024.findings-emnlp.412/",
    doi = "10.18653/v1/2024.findings-emnlp.412",
    pages = "7017--7034"
}

@article{gardner1992spelling,
     ISSN = {00493155, 1938369X},
     URL = {http://www.jstor.org/stable/43095181},
     author = {Sylvia A. Gardner},
     journal = {Technical Communication},
     number = {1},
     pages = {50--53},
     publisher = {Society for Technical Communication},
     title = {{Spelling Errors in Online Databases: What the Technical Communicator Should Know}},
     urldate = {2025-07-16},
     volume = {39},
     year = {1992}
}

@book{lisbach2013linguistic,
  author       = {Bertrand Lisbach and
                  Victoria Meyer},
  title        = {{Linguistic Identity Matching}},
  publisher    = {Springer},
  year         = {2013},
  url          = {https://doi.org/10.1007/978-3-8348-2095-2},
  doi          = {10.1007/978-3-8348-2095-2},
  isbn         = {978-3-8348-1370-1},
  bibsource    = {dblp computer science bibliography, https://dblp.org}
}

@inproceedings{gao2018black,
  author       = {Ji Gao and
                  Jack Lanchantin and
                  Mary Lou Soffa and
                  Yanjun Qi},
  title        = {{Black-Box Generation of Adversarial Text Sequences to Evade Deep Learning Classifiers}},
  booktitle    = {2018 {IEEE} Security and Privacy Workshops, {SP} Workshops 2018, San
                  Francisco, CA, USA, May 24, 2018},
  pages        = {50--56},
  publisher    = {{IEEE} Computer Society},
  year         = {2018},
  url          = {https://doi.org/10.1109/SPW.2018.00016},
  doi          = {10.1109/SPW.2018.00016},
  bibsource    = {dblp computer science bibliography, https://dblp.org}
}

@inproceedings{pruthi-etal-2019-combating,
    title = "Combating Adversarial Misspellings with Robust Word Recognition",
    author = "Pruthi, Danish  and
      Dhingra, Bhuwan  and
      Lipton, Zachary C.",
    editor = "Korhonen, Anna  and
      Traum, David  and
      M{\`a}rquez, Llu{\'i}s",
    booktitle = "Proceedings of the 57th Annual Meeting of the Association for Computational Linguistics",
    month = jul,
    year = "2019",
    address = "Florence, Italy",
    publisher = "Association for Computational Linguistics",
    url = "https://aclanthology.org/P19-1561/",
    doi = "10.18653/v1/P19-1561",
    pages = "5582--5591"
}

@inproceedings{zhang-etal-2022-interpreting,
    title = "Interpreting the Robustness of Neural {NLP} Models to Textual Perturbations",
    author = "Zhang, Yunxiang  and
      Pan, Liangming  and
      Tan, Samson  and
      Kan, Min-Yen",
    editor = "Muresan, Smaranda  and
      Nakov, Preslav  and
      Villavicencio, Aline",
    booktitle = "Findings of the Association for Computational Linguistics: ACL 2022",
    month = may,
    year = "2022",
    address = "Dublin, Ireland",
    publisher = "Association for Computational Linguistics",
    url = "https://aclanthology.org/2022.findings-acl.315/",
    doi = "10.18653/v1/2022.findings-acl.315",
    pages = "3993--4007"
}

@inproceedings{li2019textbugger,
  author       = {Jinfeng Li and
                  Shouling Ji and
                  Tianyu Du and
                  Bo Li and
                  Ting Wang},
  title        = {TextBugger: Generating Adversarial Text Against Real-world Applications},
  booktitle    = {26th Annual Network and Distributed System Security Symposium, {NDSS}
                  2019, San Diego, California, USA, February 24-27, 2019},
  publisher    = {The Internet Society},
  year         = {2019},
  url          = {https://www.ndss-symposium.org/ndss-paper/textbugger-generating-adversarial-text-against-real-world-applications/},
  bibsource    = {dblp computer science bibliography, https://dblp.org}
}

@inproceedings{garg-ramakrishnan-2020-bae,
    title = "{BAE}: {BERT}-based Adversarial Examples for Text Classification",
    author = "Garg, Siddhant  and
      Ramakrishnan, Goutham",
    editor = "Webber, Bonnie  and
      Cohn, Trevor  and
      He, Yulan  and
      Liu, Yang",
    booktitle = "Proceedings of the 2020 Conference on Empirical Methods in Natural Language Processing (EMNLP)",
    month = nov,
    year = "2020",
    address = "Online",
    publisher = "Association for Computational Linguistics",
    url = "https://aclanthology.org/2020.emnlp-main.498/",
    doi = "10.18653/v1/2020.emnlp-main.498",
    pages = "6174--6181"
}

@inproceedings{moradi-samwald-2021-evaluating,
    title = "Evaluating the Robustness of Neural Language Models to Input Perturbations",
    author = "Moradi, Milad  and
      Samwald, Matthias",
    editor = "Moens, Marie-Francine  and
      Huang, Xuanjing  and
      Specia, Lucia  and
      Yih, Scott Wen-tau",
    booktitle = "Proceedings of the 2021 Conference on Empirical Methods in Natural Language Processing",
    month = nov,
    year = "2021",
    address = "Online and Punta Cana, Dominican Republic",
    publisher = "Association for Computational Linguistics",
    url = "https://aclanthology.org/2021.emnlp-main.117/",
    doi = "10.18653/v1/2021.emnlp-main.117",
    pages = "1558--1570"
}

@inproceedings{jin2020bert,
  author       = {Di Jin and
                  Zhijing Jin and
                  Joey Tianyi Zhou and
                  Peter Szolovits},
  title        = {Is {BERT} Really Robust? {A} Strong Baseline for Natural Language
                  Attack on Text Classification and Entailment},
  booktitle    = {The Thirty-Fourth {AAAI} Conference on Artificial Intelligence, {AAAI}
                  2020, The Thirty-Second Innovative Applications of Artificial Intelligence
                  Conference, {IAAI} 2020, The Tenth {AAAI} Symposium on Educational
                  Advances in Artificial Intelligence, {EAAI} 2020, New York, NY, USA,
                  February 7-12, 2020},
  pages        = {8018--8025},
  publisher    = {{AAAI} Press},
  year         = {2020},
  url          = {https://doi.org/10.1609/aaai.v34i05.6311},
  doi          = {10.1609/AAAI.V34I05.6311},
  bibsource    = {dblp computer science bibliography, https://dblp.org}
}

@inproceedings{zhou2024mathattack,
  author       = {Zihao Zhou and
                  Qiufeng Wang and
                  Mingyu Jin and
                  Jie Yao and
                  Jianan Ye and
                  Wei Liu and
                  Wei Wang and
                  Xiaowei Huang and
                  Kaizhu Huang},
  editor       = {Michael J. Wooldridge and
                  Jennifer G. Dy and
                  Sriraam Natarajan},
  title        = {MathAttack: Attacking Large Language Models towards Math Solving Ability},
  booktitle    = {Thirty-Eighth {AAAI} Conference on Artificial Intelligence, {AAAI}
                  2024, Thirty-Sixth Conference on Innovative Applications of Artificial
                  Intelligence, {IAAI} 2024, Fourteenth Symposium on Educational Advances
                  in Artificial Intelligence, {EAAI} 2014, February 20-27, 2024, Vancouver,
                  Canada},
  pages        = {19750--19758},
  publisher    = {{AAAI} Press},
  year         = {2024},
  url          = {https://doi.org/10.1609/aaai.v38i17.29949},
  doi          = {10.1609/AAAI.V38I17.29949},
  bibsource    = {dblp computer science bibliography, https://dblp.org}
}

@inproceedings{shi2023large,
  author       = {Freda Shi and
                  Xinyun Chen and
                  Kanishka Misra and
                  Nathan Scales and
                  David Dohan and
                  Ed H. Chi and
                  Nathanael Sch{\"{a}}rli and
                  Denny Zhou},
  editor       = {Andreas Krause and
                  Emma Brunskill and
                  Kyunghyun Cho and
                  Barbara Engelhardt and
                  Sivan Sabato and
                  Jonathan Scarlett},
  title        = {Large Language Models Can Be Easily Distracted by Irrelevant Context},
  booktitle    = {International Conference on Machine Learning, {ICML} 2023, 23-29 July
                  2023, Honolulu, Hawaii, {USA}},
  series       = {Proceedings of Machine Learning Research},
  volume       = {202},
  pages        = {31210--31227},
  publisher    = {{PMLR}},
  year         = {2023},
  url          = {https://proceedings.mlr.press/v202/shi23a.html},
  bibsource    = {dblp computer science bibliography, https://dblp.org}
}

@inproceedings{xu2024llm,
  author       = {Xilie Xu and
                  Keyi Kong and
                  Ning Liu and
                  Lizhen Cui and
                  Di Wang and
                  Jingfeng Zhang and
                  Mohan S. Kankanhalli},
  title        = {An {LLM} can Fool Itself: {A} Prompt-Based Adversarial Attack},
  booktitle    = {The Twelfth International Conference on Learning Representations,
                  {ICLR} 2024, Vienna, Austria, May 7-11, 2024},
  publisher    = {OpenReview.net},
  year         = {2024},
  url          = {https://openreview.net/forum?id=VVgGbB9TNV},
  bibsource    = {dblp computer science bibliography, https://dblp.org}
}

@misc{lanham2023measuring,
      title={Measuring Faithfulness in Chain-of-Thought Reasoning}, 
      author={Tamera Lanham and Anna Chen and Ansh Radhakrishnan and Benoit Steiner and Carson Denison and Danny Hernandez and Dustin Li and Esin Durmus and Evan Hubinger and Jackson Kernion and Kamilė Lukošiūtė and Karina Nguyen and Newton Cheng and Nicholas Joseph and Nicholas Schiefer and Oliver Rausch and Robin Larson and Sam McCandlish and Sandipan Kundu and Saurav Kadavath and Shannon Yang and Thomas Henighan and Timothy Maxwell and Timothy Telleen-Lawton and Tristan Hume and Zac Hatfield-Dodds and Jared Kaplan and Jan Brauner and Samuel R. Bowman and Ethan Perez},
      year={2023},
      eprint={2307.13702},
      archivePrefix={arXiv},
      primaryClass={cs.AI},
      url={https://arxiv.org/abs/2307.13702}, 
}

@article{logan2016different,
  title={Different (key) strokes for different folks: How standard and nonstandard typists balance Fitts’ law and Hick’s law.},
  author={Logan, Gordon D and Ulrich, Jana E and Lindsey, Dakota RB},
  journal={Journal of Experimental Psychology: Human Perception and Performance},
  volume={42},
  number={12},
  pages={2084},
  year={2016},
  publisher={American Psychological Association},
  url={https://psycnet.apa.org/record/2016-49569-001}
}

@misc{dam2024completesurveyllmbasedai,
      title={A Complete Survey on LLM-based AI Chatbots}, 
      author={Sumit Kumar Dam and Choong Seon Hong and Yu Qiao and Chaoning Zhang},
      year={2024},
      eprint={2406.16937},
      archivePrefix={arXiv},
      primaryClass={cs.CL},
      url={https://arxiv.org/abs/2406.16937}, 
}

@misc{naveed2024comprehensiveoverviewlargelanguage,
      title={A Comprehensive Overview of Large Language Models}, 
      author={Humza Naveed and Asad Ullah Khan and Shi Qiu and Muhammad Saqib and Saeed Anwar and Muhammad Usman and Naveed Akhtar and Nick Barnes and Ajmal Mian},
      year={2024},
      eprint={2307.06435},
      archivePrefix={arXiv},
      primaryClass={cs.CL},
      url={https://arxiv.org/abs/2307.06435}, 
}

@article{raza2025industrial,
  title={Industrial applications of large language models},
  author={Raza, Mubashar and Jahangir, Zarmina and Riaz, Muhammad Bilal and Saeed, Muhammad Jasim and Sattar, Muhammad Awais},
  journal={Scientific Reports},
  volume={15},
  number={1},
  pages={13755},
  year={2025},
  publisher={Nature Publishing Group UK London},
  url={https://doi.org/10.1038/s41598-025-98483-1}
}

@incollection{wengelin2007word,
  title={The word-level focus in text production by adults with reading and writing difficulties},
  author={Wengelin, {\AA}sa},
  booktitle={Writing and cognition},
  pages={67--82},
  year={2007},
  publisher={Brill},
  url={https://doi.org/10.1163/9781849508223_006}
}

@article{conijn2019typo,
  title={How to typo? Building a process-based model of typographic error revisions},
  author={Conijn, Rianne and Van Zaanen, Menno and Leijten, Mari{\"e}lle and Van Waes, Luuk},
  journal={Journal of Writing Analytics},
  volume={3},
  pages={69--95},
  year={2019},
  publisher={Colorado State University Open Press},
  url={https://wac.colostate.edu/docs/jwa/vol3/conijin.pdf}
}

@misc{sun2020advbertbertrobustmisspellings,
      title={Adv-BERT: BERT is not robust on misspellings! Generating nature adversarial samples on BERT}, 
      author={Lichao Sun and Kazuma Hashimoto and Wenpeng Yin and Akari Asai and Jia Li and Philip Yu and Caiming Xiong},
      year={2020},
      eprint={2003.04985},
      archivePrefix={arXiv},
      primaryClass={cs.CL},
      url={https://arxiv.org/abs/2003.04985}, 
}

@inproceedings{shi2025simulating,
  author       = {Danqing Shi and
                  Yujun Zhu and
                  Francisco Erivaldo Fernandes Junior and
                  Shumin Zhai and
                  Antti Oulasvirta},
  editor       = {Naomi Yamashita and
                  Vanessa Evers and
                  Koji Yatani and
                  Sharon Xianghua Ding and
                  Bongshin Lee and
                  Marshini Chetty and
                  Phoebe O. Toups Dugas},
  title        = {Simulating Errors in Touchscreen Typing},
  booktitle    = {Proceedings of the 2025 {CHI} Conference on Human Factors in Computing
                  Systems, {CHI} 2025, YokohamaJapan, 26 April 2025- 1 May 2025},
  pages        = {1086:1--1086:13},
  publisher    = {{ACM}},
  year         = {2025},
  url          = {https://doi.org/10.1145/3706598.3713153},
  doi          = {10.1145/3706598.3713153},
  bibsource    = {dblp computer science bibliography, https://dblp.org}
}

@inproceedings{jussi2021Touchscreen,
  author       = {Jussi Jokinen and
                  Aditya Acharya and
                  Mohammad Uzair and
                  Xinhui Jiang and
                  Antti Oulasvirta},
  editor       = {Yoshifumi Kitamura and
                  Aaron Quigley and
                  Katherine Isbister and
                  Takeo Igarashi and
                  Pernille Bj{\o}rn and
                  Steven Mark Drucker},
  title        = {Touchscreen Typing As Optimal Supervisory Control},
  booktitle    = {{CHI} '21: {CHI} Conference on Human Factors in Computing Systems,
                  Virtual Event / Yokohama, Japan, May 8-13, 2021},
  pages        = {720:1--720:14},
  publisher    = {{ACM}},
  year         = {2021},
  url          = {https://doi.org/10.1145/3411764.3445483},
  doi          = {10.1145/3411764.3445483},
  bibsource    = {dblp computer science bibliography, https://dblp.org}
}

@inproceedings{ribeiro-etal-2020-beyond,
    title = "Beyond Accuracy: Behavioral Testing of {NLP} Models with {C}heck{L}ist",
    author = "Ribeiro, Marco Tulio  and
      Wu, Tongshuang  and
      Guestrin, Carlos  and
      Singh, Sameer",
    editor = "Jurafsky, Dan  and
      Chai, Joyce  and
      Schluter, Natalie  and
      Tetreault, Joel",
    booktitle = "Proceedings of the 58th Annual Meeting of the Association for Computational Linguistics",
    month = jul,
    year = "2020",
    address = "Online",
    publisher = "Association for Computational Linguistics",
    url = "https://aclanthology.org/2020.acl-main.442/",
    doi = "10.18653/v1/2020.acl-main.442",
    pages = "4902--4912"
}

@inproceedings{wang-etal-2024-resilience,
    title = "Resilience of Large Language Models for Noisy Instructions",
    author = "Wang, Bin  and
      Wei, Chengwei  and
      Liu, Zhengyuan  and
      Lin, Geyu  and
      Chen, Nancy F.",
    editor = "Al-Onaizan, Yaser  and
      Bansal, Mohit  and
      Chen, Yun-Nung",
    booktitle = "Findings of the Association for Computational Linguistics: EMNLP 2024",
    month = nov,
    year = "2024",
    address = "Miami, Florida, USA",
    publisher = "Association for Computational Linguistics",
    url = "https://aclanthology.org/2024.findings-emnlp.697/",
    doi = "10.18653/v1/2024.findings-emnlp.697",
    pages = "11939--11950"
}

@misc{wang2023robustnesschatgptadversarialoutofdistribution,
      title={On the Robustness of ChatGPT: An Adversarial and Out-of-distribution Perspective}, 
      author={Jindong Wang and Xixu Hu and Wenxin Hou and Hao Chen and Runkai Zheng and Yidong Wang and Linyi Yang and Haojun Huang and Wei Ye and Xiubo Geng and Binxin Jiao and Yue Zhang and Xing Xie},
      year={2023},
      eprint={2302.12095},
      archivePrefix={arXiv},
      primaryClass={cs.AI},
      url={https://arxiv.org/abs/2302.12095}, 
}

@inproceedings{zhu2024promptrobust,
  author       = {Kaijie Zhu and
                  Jindong Wang and
                  Jiaheng Zhou and
                  Zichen Wang and
                  Hao Chen and
                  Yidong Wang and
                  Linyi Yang and
                  Wei Ye and
                  Yue Zhang and
                  Neil Gong and
                  Xing Xie},
  editor       = {Bo Li and
                  Wenyuan Xu and
                  Jieshan Chen and
                  Yang Zhang and
                  Jason Xue and
                  Shuo Wang and
                  Guangdong Bai and
                  Xingliang Yuan},
  title        = {PromptRobust: Towards Evaluating the Robustness of Large Language
                  Models on Adversarial Prompts},
  booktitle    = {Proceedings of the 1st {ACM} Workshop on Large {AI} Systems and Models
                  with Privacy and Safety Analysis, {LAMPS} 2024, Salt Lake City, UT,
                  USA, October 14-18, 2024},
  pages        = {57--68},
  publisher    = {{ACM}},
  year         = {2024},
  url          = {https://doi.org/10.1145/3689217.3690621},
  doi          = {10.1145/3689217.3690621},
  bibsource    = {dblp computer science bibliography, https://dblp.org}
}

@misc{zhang2025evaluatingimprovingrobustnesslarge,
      title={Evaluating and Improving Robustness in Large Language Models: A Survey and Future Directions}, 
      author={Kun Zhang and Le Wu and Kui Yu and Guangyi Lv and Dacao Zhang},
      year={2025},
      eprint={2506.11111},
      archivePrefix={arXiv},
      primaryClass={cs.CL},
      url={https://arxiv.org/abs/2506.11111}, 
}

@inproceedings{cooper-stickland-etal-2023-robustification,
    title = "Robustification of Multilingual Language Models to Real-world Noise in Crosslingual Zero-shot Settings with Robust Contrastive Pretraining",
    author = "Cooper Stickland, Asa  and
      Sengupta, Sailik  and
      Krone, Jason  and
      Mansour, Saab  and
      He, He",
    editor = "Vlachos, Andreas  and
      Augenstein, Isabelle",
    booktitle = "Proceedings of the 17th Conference of the European Chapter of the Association for Computational Linguistics",
    month = may,
    year = "2023",
    address = "Dubrovnik, Croatia",
    publisher = "Association for Computational Linguistics",
    url = "https://aclanthology.org/2023.eacl-main.100/",
    doi = "10.18653/v1/2023.eacl-main.100",
    pages = "1375--1391"
}

@misc{aliakbarzadeh2025exploringrobustnessmultilingualllms,
      title={Exploring Robustness of Multilingual LLMs on Real-World Noisy Data}, 
      author={Amirhossein Aliakbarzadeh and Lucie Flek and Akbar Karimi},
      year={2025},
      eprint={2501.08322},
      archivePrefix={arXiv},
      primaryClass={cs.CL},
      url={https://arxiv.org/abs/2501.08322}, 
}

@inproceedings{devlin-etal-2019-bert,
    title = "{BERT}: Pre-training of Deep Bidirectional Transformers for Language Understanding",
    author = "Devlin, Jacob  and
      Chang, Ming-Wei  and
      Lee, Kenton  and
      Toutanova, Kristina",
    editor = "Burstein, Jill  and
      Doran, Christy  and
      Solorio, Thamar",
    booktitle = "Proceedings of the 2019 Conference of the North {A}merican Chapter of the Association for Computational Linguistics: Human Language Technologies, Volume 1 (Long and Short Papers)",
    month = jun,
    year = "2019",
    address = "Minneapolis, Minnesota",
    publisher = "Association for Computational Linguistics",
    url = "https://aclanthology.org/N19-1423",
    doi = "10.18653/v1/N19-1423",
    pages = "4171--4186"
}

@inproceedings{conneau-etal-2020-unsupervised,
    title = "Unsupervised Cross-lingual Representation Learning at Scale",
    author = "Conneau, Alexis  and
      Khandelwal, Kartikay  and
      Goyal, Naman  and
      Chaudhary, Vishrav  and
      Wenzek, Guillaume  and
      Guzm{\'a}n, Francisco  and
      Grave, Edouard  and
      Ott, Myle  and
      Zettlemoyer, Luke  and
      Stoyanov, Veselin",
    booktitle = "Proceedings of the 58th Annual Meeting of the Association for Computational Linguistics",
    month = jul,
    year = "2020",
    address = "Online",
    publisher = "Association for Computational Linguistics",
    url = "https://aclanthology.org/2020.acl-main.747",
    doi = "10.18653/v1/2020.acl-main.747",
    pages = "8440--8451",
}

@misc{yang2025qwen3technicalreport,
      title={Qwen3 Technical Report}, 
      author={An Yang and Anfeng Li and Baosong Yang and Beichen Zhang and Binyuan Hui and Bo Zheng and Bowen Yu and Chang Gao and Chengen Huang and Chenxu Lv and Chujie Zheng and Dayiheng Liu and Fan Zhou and Fei Huang and Feng Hu and Hao Ge and Haoran Wei and Huan Lin and Jialong Tang and Jian Yang and Jianhong Tu and Jianwei Zhang and Jianxin Yang and Jiaxi Yang and Jing Zhou and Jingren Zhou and Junyang Lin and Kai Dang and Keqin Bao and Kexin Yang and Le Yu and Lianghao Deng and Mei Li and Mingfeng Xue and Mingze Li and Pei Zhang and Peng Wang and Qin Zhu and Rui Men and Ruize Gao and Shixuan Liu and Shuang Luo and Tianhao Li and Tianyi Tang and Wenbiao Yin and Xingzhang Ren and Xinyu Wang and Xinyu Zhang and Xuancheng Ren and Yang Fan and Yang Su and Yichang Zhang and Yinger Zhang and Yu Wan and Yuqiong Liu and Zekun Wang and Zeyu Cui and Zhenru Zhang and Zhipeng Zhou and Zihan Qiu},
      year={2025},
      eprint={2505.09388},
      archivePrefix={arXiv},
      primaryClass={cs.CL},
      url={https://arxiv.org/abs/2505.09388}, 
}

@misc{gemmateam2025gemma3technicalreport,
      title={Gemma 3 Technical Report}, 
      author={{Gemma Team} and Aishwarya Kamath and Johan Ferret and Shreya Pathak and Nino Vieillard and Ramona Merhej and Sarah Perrin and Tatiana Matejovicova and Alexandre Ramé and Morgane Rivière and Louis Rouillard and Thomas Mesnard and Geoffrey Cideron and Jean-bastien Grill and Sabela Ramos and Edouard Yvinec and Michelle Casbon and Etienne Pot and Ivo Penchev and Gaël Liu and Francesco Visin and Kathleen Kenealy and Lucas Beyer and Xiaohai Zhai and Anton Tsitsulin and Robert Busa-Fekete and Alex Feng and Noveen Sachdeva and Benjamin Coleman and Yi Gao and Basil Mustafa and Iain Barr and Emilio Parisotto and David Tian and Matan Eyal and Colin Cherry and Jan-Thorsten Peter and Danila Sinopalnikov and Surya Bhupatiraju and Rishabh Agarwal and Mehran Kazemi and Dan Malkin and Ravin Kumar and Others},
      year={2025},
      eprint={2503.19786},
      archivePrefix={arXiv},
      primaryClass={cs.CL},
      url={https://arxiv.org/abs/2503.19786}, 
}

@misc{olmo20252olmo2furious,
      title={2 OLMo 2 Furious}, 
      author={{Team OLMo} and Pete Walsh and Luca Soldaini and Dirk Groeneveld and Kyle Lo and Shane Arora and Akshita Bhagia and Yuling Gu and Shengyi Huang and Matt Jordan and Nathan Lambert and Dustin Schwenk and Oyvind Tafjord and Taira Anderson and David Atkinson and Faeze Brahman and Christopher Clark and Pradeep Dasigi and Nouha Dziri and Michal Guerquin and Hamish Ivison and Pang Wei Koh and Jiacheng Liu and Saumya Malik and William Merrill and Lester James V. Miranda and Jacob Morrison and Tyler Murray and Crystal Nam and Valentina Pyatkin and Aman Rangapur and Michael Schmitz and Sam Skjonsberg and David Wadden and Christopher Wilhelm and Michael Wilson and Luke Zettlemoyer and Ali Farhadi and Noah A. Smith and Hannaneh Hajishirzi},
      year={2025},
      eprint={2501.00656},
      archivePrefix={arXiv},
      primaryClass={cs.CL},
      url={https://arxiv.org/abs/2501.00656}, 
}

@inproceedings{conneau-etal-2018-xnli,
    title = "{XNLI}: Evaluating Cross-lingual Sentence Representations",
    author = "Conneau, Alexis  and
      Rinott, Ruty  and
      Lample, Guillaume  and
      Williams, Adina  and
      Bowman, Samuel  and
      Schwenk, Holger  and
      Stoyanov, Veselin",
    editor = "Riloff, Ellen  and
      Chiang, David  and
      Hockenmaier, Julia  and
      Tsujii, Jun{'}ichi",
    booktitle = "Proceedings of the 2018 Conference on Empirical Methods in Natural Language Processing",
    month = oct # "-" # nov,
    year = "2018",
    address = "Brussels, Belgium",
    publisher = "Association for Computational Linguistics",
    url = "https://aclanthology.org/D18-1269/",
    doi = "10.18653/v1/D18-1269",
    pages = "2475--2485"
}

@inproceedings{bandarkar-etal-2024-belebele,
    title = "The Belebele Benchmark: a Parallel Reading Comprehension Dataset in 122 Language Variants",
    author = "Bandarkar, Lucas  and
      Liang, Davis  and
      Muller, Benjamin  and
      Artetxe, Mikel  and
      Shukla, Satya Narayan  and
      Husa, Donald  and
      Goyal, Naman  and
      Krishnan, Abhinandan  and
      Zettlemoyer, Luke  and
      Khabsa, Madian",
    editor = "Ku, Lun-Wei  and
      Martins, Andre  and
      Srikumar, Vivek",
    booktitle = "Proceedings of the 62nd Annual Meeting of the Association for Computational Linguistics (Volume 1: Long Papers)",
    month = aug,
    year = "2024",
    address = "Bangkok, Thailand",
    publisher = "Association for Computational Linguistics",
    url = "https://aclanthology.org/2024.acl-long.44/",
    doi = "10.18653/v1/2024.acl-long.44",
    pages = "749--775"
}

@inproceedings{mmmlu,
  author       = {Dan Hendrycks and
                  Collin Burns and
                  Steven Basart and
                  Andy Zou and
                  Mantas Mazeika and
                  Dawn Song and
                  Jacob Steinhardt},
  title        = {Measuring Massive Multitask Language Understanding},
  booktitle    = {9th International Conference on Learning Representations, {ICLR} 2021,
                  Virtual Event, Austria, May 3-7, 2021},
  publisher    = {OpenReview.net},
  year         = {2021},
  url          = {https://openreview.net/forum?id=d7KBjmI3GmQ},
  bibsource    = {dblp computer science bibliography, https://dblp.org}
}

@inproceedings{mgsm,
  author       = {Freda Shi and
                  Mirac Suzgun and
                  Markus Freitag and
                  Xuezhi Wang and
                  Suraj Srivats and
                  Soroush Vosoughi and
                  Hyung Won Chung and
                  Yi Tay and
                  Sebastian Ruder and
                  Denny Zhou and
                  Dipanjan Das and
                  Jason Wei},
  title        = {{Language models are multilingual chain-of-thought reasoners}},
  booktitle    = {The Eleventh International Conference on Learning Representations,
                  {ICLR} 2023, Kigali, Rwanda, May 1-5, 2023},
  publisher    = {OpenReview.net},
  year         = {2023},
  url          = {https://openreview.net/forum?id=fR3wGCk-IXp},
  bibsource    = {dblp computer science bibliography, https://dblp.org}
}

@article{GSM8K,
  author       = {Karl Cobbe and
                  Vineet Kosaraju and
                  Mohammad Bavarian and
                  Mark Chen and
                  Heewoo Jun and
                  Lukasz Kaiser and
                  Matthias Plappert and
                  Jerry Tworek and
                  Jacob Hilton and
                  Reiichiro Nakano and
                  Christopher Hesse and
                  John Schulman},
  title        = {{Training Verifiers to Solve Math Word Problems}},
  journal      = {CoRR},
  volume       = {abs/2110.14168},
  year         = {2021},
  url          = {https://arxiv.org/abs/2110.14168},
  eprinttype    = {arXiv},
  eprint       = {2110.14168},
  bibsource    = {dblp computer science bibliography, https://dblp.org}
}

@misc{arabic-gsm8k,
  title={{Arabic GSM8K: Arabic Grade School Math Dataset}},
  author={Omartificial-Intelligence-Space},
  year={2025},
  howpublished={\url{https://huggingface.co/datasets/Omartificial-Intelligence-Space/Arabic-gsm8k}}
}

@misc{hindi-gsm8k,
      title={{Towards Inducing Document-Level Abilities in Standard Multilingual Neural Machine Translation Models}}, 
      author={Varun Gumma and Pranjal A. Chitale and Kalika Bali},
      year={2024},
      eprint={2408.11382},
      archivePrefix={arXiv},
      primaryClass={cs.CL},
      url={https://arxiv.org/abs/2408.11382}, 
}

@misc{flores200,
      title={No Language Left Behind: Scaling Human-Centered Machine Translation}, 
      author={{NLLB Team} and Marta R. Costa-jussà and James Cross and Onur Çelebi and Maha Elbayad and Kenneth Heafield and Kevin Heffernan and Elahe Kalbassi and Janice Lam and Daniel Licht and Jean Maillard and Anna Sun and Skyler Wang and Guillaume Wenzek and Al Youngblood and Bapi Akula and Loic Barrault and Gabriel Mejia Gonzalez and Prangthip Hansanti and John Hoffman and Semarley Jarrett and Kaushik Ram Sadagopan and Dirk Rowe and Shannon Spruit and Chau Tran and Pierre Andrews and Necip Fazil Ayan and Shruti Bhosale and Sergey Edunov and Angela Fan and Cynthia Gao and Vedanuj Goswami and Francisco Guzmán and Philipp Koehn and Alexandre Mourachko and Christophe Ropers and Safiyyah Saleem and Holger Schwenk and Jeff Wang},
      year={2022},
      eprint={2207.04672},
      archivePrefix={arXiv},
      primaryClass={cs.CL},
      url={https://arxiv.org/abs/2207.04672}, 
}

@article{wang2025perturbations,
  author       = {Haoyu Wang and
                  Guozheng Ma and
                  Cong Yu and
                  Ning Gui and
                  Linrui Zhang and
                  Zhiqi Huang and
                  Suwei Ma and
                  Yongzhe Chang and
                  Sen Zhang and
                  Li Shen and
                  Xueqian Wang and
                  Peilin Zhao and
                  Dacheng Tao},
  title        = {Are Large Language Models Really Robust to Word-Level Perturbations?},
  journal      = {Trans. Mach. Learn. Res.},
  volume       = {2025},
  year         = {2025},
  url          = {https://openreview.net/forum?id=rWSiBknwQa},
  bibsource    = {dblp computer science bibliography, https://dblp.org}
}

@article{Won2o24scaling,
  author       = {Hyung Won Chung and
                  Le Hou and
                  Shayne Longpre and
                  Barret Zoph and
                  Yi Tay and
                  William Fedus and
                  Yunxuan Li and
                  Xuezhi Wang and
                  Mostafa Dehghani and
                  Siddhartha Brahma and
                  Albert Webson and
                  Shixiang Shane Gu and
                  Zhuyun Dai and
                  Mirac Suzgun and
                  Xinyun Chen and
                  Aakanksha Chowdhery and
                  Alex Castro{-}Ros and
                  Marie Pellat and
                  Kevin Robinson and
                  Dasha Valter and
                  Sharan Narang and
                  Gaurav Mishra and
                  Adams Yu and
                  Vincent Y. Zhao and
                  Yanping Huang and
                  Andrew M. Dai and
                  Hongkun Yu and
                  Slav Petrov and
                  Ed H. Chi and
                  Jeff Dean and
                  Jacob Devlin and
                  Adam Roberts and
                  Denny Zhou and
                  Quoc V. Le and
                  Jason Wei},
  title        = {Scaling Instruction-Finetuned Language Models},
  journal      = {J. Mach. Learn. Res.},
  volume       = {25},
  pages        = {70:1--70:53},
  year         = {2024},
  url          = {https://jmlr.org/papers/v25/23-0870.html},
  bibsource    = {dblp computer science bibliography, https://dblp.org}
}

@inproceedings{liu-etal-2023-logicot,
    title = "{L}ogi{C}o{T}: Logical Chain-of-Thought Instruction Tuning",
    author = "Liu, Hanmeng  and
      Teng, Zhiyang  and
      Cui, Leyang  and
      Zhang, Chaoli  and
      Zhou, Qiji  and
      Zhang, Yue",
    editor = "Bouamor, Houda  and
      Pino, Juan  and
      Bali, Kalika",
    booktitle = "Findings of the Association for Computational Linguistics: EMNLP 2023",
    month = dec,
    year = "2023",
    address = "Singapore",
    publisher = "Association for Computational Linguistics",
    url = "https://aclanthology.org/2023.findings-emnlp.191/",
    doi = "10.18653/v1/2023.findings-emnlp.191",
    pages = "2908--2921"
}

@inproceedings{Brown2020fewshot,
  author       = {Tom B. Brown and
                  Benjamin Mann and
                  Nick Ryder and
                  Melanie Subbiah and
                  Jared Kaplan and
                  Prafulla Dhariwal and
                  Arvind Neelakantan and
                  Pranav Shyam and
                  Girish Sastry and
                  Amanda Askell and
                  Sandhini Agarwal and
                  Ariel Herbert{-}Voss and
                  Gretchen Krueger and
                  Tom Henighan and
                  Rewon Child and
                  Aditya Ramesh and
                  Daniel M. Ziegler and
                  Jeffrey Wu and
                  Clemens Winter and
                  Christopher Hesse and
                  Mark Chen and
                  Eric Sigler and
                  Mateusz Litwin and
                  Scott Gray and
                  Benjamin Chess and
                  Jack Clark and
                  Christopher Berner and
                  Sam McCandlish and
                  Alec Radford and
                  Ilya Sutskever and
                  Dario Amodei},
  editor       = {Hugo Larochelle and
                  Marc'Aurelio Ranzato and
                  Raia Hadsell and
                  Maria{-}Florina Balcan and
                  Hsuan{-}Tien Lin},
  title        = {Language Models are Few-Shot Learners},
  booktitle    = {Advances in Neural Information Processing Systems 33: Annual Conference
                  on Neural Information Processing Systems 2020, NeurIPS 2020, December
                  6-12, 2020, virtual},
  year         = {2020},
  url          = {https://proceedings.neurips.cc/paper/2020/hash/1457c0d6bfcb4967418bfb8ac142f64a-Abstract.html},
  bibsource    = {dblp computer science bibliography, https://dblp.org}
}

@article{schick-schutze-2022-true,
    title = "True Few-Shot Learning with {P}rompts{---}{A} Real-World Perspective",
    author = {Schick, Timo  and
      Sch{\"u}tze, Hinrich},
    editor = "Roark, Brian  and
      Nenkova, Ani",
    journal = "Transactions of the Association for Computational Linguistics",
    volume = "10",
    year = "2022",
    address = "Cambridge, MA",
    publisher = "MIT Press",
    url = "https://aclanthology.org/2022.tacl-1.41/",
    doi = "10.1162/tacl_a_00485",
    pages = "716--731"
}

@inproceedings{baba-suzuki-2012-spelling,
    title = "How Are Spelling Errors Generated and Corrected? A Study of Corrected and Uncorrected Spelling Errors Using Keystroke Logs",
    author = "Baba, Yukino  and
      Suzuki, Hisami",
    editor = "Li, Haizhou  and
      Lin, Chin-Yew  and
      Osborne, Miles  and
      Lee, Gary Geunbae  and
      Park, Jong C.",
    booktitle = "Proceedings of the 50th Annual Meeting of the Association for Computational Linguistics (Volume 2: Short Papers)",
    month = jul,
    year = "2012",
    address = "Jeju Island, Korea",
    publisher = "Association for Computational Linguistics",
    url = "https://aclanthology.org/P12-2073/",
    pages = "373--377"
}

@article{macneilage1964typing,
  title={Typing errors as clues to serial ordering mechanisms in language behaviour},
  author={MacNeilage, Peter F},
  journal={Language and speech},
  volume={7},
  number={3},
  pages={144--159},
  year={1964},
  publisher={SAGE Publications Sage UK: London, England},
  url={https://pure.mpg.de/rest/items/item_2355504_3/component/file_2355503/content}
}

@inproceedings{Hedderich2021Label,
  author       = {Michael A. Hedderich and
                  Jonas Fischer and
                  Dietrich Klakow and
                  Jilles Vreeken},
  editor       = {Kamalika Chaudhuri and
                  Stefanie Jegelka and
                  Le Song and
                  Csaba Szepesv{\'{a}}ri and
                  Gang Niu and
                  Sivan Sabato},
  title        = {Label-Descriptive Patterns and Their Application to Characterizing
                  Classification Errors},
  booktitle    = {International Conference on Machine Learning, {ICML} 2022, 17-23 July
                  2022, Baltimore, Maryland, {USA}},
  series       = {Proceedings of Machine Learning Research},
  volume       = {162},
  pages        = {8691--8707},
  publisher    = {{PMLR}},
  year         = {2022},
  url          = {https://proceedings.mlr.press/v162/hedderich22a.html},
  bibsource    = {dblp computer science bibliography, https://dblp.org}
}

@article{peterson1986note,
  title={A note on undetected typing errors},
  author={Peterson, James L},
  journal={Communications of the ACM},
  volume={29},
  number={7},
  pages={633--637},
  year={1986},
  publisher={ACM New York, NY, USA},
  url={https://dl.acm.org/doi/pdf/10.1145/6138.6146}
}

@article{kukich1992techniques,
  title={Techniques for automatically correcting words in text},
  author={Kukich, Karen},
  journal={ACM computing surveys (CSUR)},
  volume={24},
  number={4},
  pages={377--439},
  year={1992},
  publisher={ACM New York, NY, USA},
  url={https://dl.acm.org/doi/10.1145/146370.146380}
}

@inproceedings{qi-etal-2025-models,
    title = "When Models Reason in Your Language: Controlling Thinking Language Comes at the Cost of Accuracy",
    author = "Qi, Jirui  and
      Chen, Shan  and
      Xiong, Zidi  and
      Fern{\'a}ndez, Raquel  and
      Bitterman, Danielle  and
      Bisazza, Arianna",
    editor = "Christodoulopoulos, Christos  and
      Chakraborty, Tanmoy  and
      Rose, Carolyn  and
      Peng, Violet",
    booktitle = "Findings of the Association for Computational Linguistics: EMNLP 2025",
    month = nov,
    year = "2025",
    address = "Suzhou, China",
    publisher = "Association for Computational Linguistics",
    url = "https://aclanthology.org/2025.findings-emnlp.1103/",
    doi = "10.18653/v1/2025.findings-emnlp.1103",
    pages = "20279--20296"
}

@inproceedings{schmidtova-etal-2026-important,
    title = "How Important is `Perfect' {E}nglish for Machine Translation Prompts?",
    author = {Schmidtov{\'a}, Patr{\'i}cia  and
      Bafna, Niyati  and
      Aycock, Seth  and
      Vico, Gianluca  and
      Kamzela, Wiktor  and
      H{\"a}mmerl, Kathy  and
      Zouhar, Vil{\'e}m},
    editor = "Demberg, Vera  and
      Inui, Kentaro  and
      Marquez, Llu{\'i}s",
    booktitle = "Findings of the {A}ssociation for {C}omputational {L}inguistics: {EACL} 2026",
    month = mar,
    year = "2026",
    address = "Rabat, Morocco",
    publisher = "Association for Computational Linguistics",
    url = "https://aclanthology.org/2026.findings-eacl.38/",
    doi = "10.18653/v1/2026.findings-eacl.38",
    pages = "760--777"
}

@inproceedings{zhao-etal-2026-comprehensive,
    title = "A Comprehensive Evaluation of Multilingual Chain-of-Thought Reasoning: Performance, Consistency, and Faithfulness Across Languages",
    author = "Zhao, Raoyuan  and
      Liu, Yihong  and
      Schuetze, Hinrich  and
      Hedderich, Michael A.",
    editor = "Demberg, Vera  and
      Inui, Kentaro  and
      Marquez, Llu{\'i}s",
    booktitle = "Findings of the {A}ssociation for {C}omputational {L}inguistics: {EACL} 2026",
    month = mar,
    year = "2026",
    address = "Rabat, Morocco",
    publisher = "Association for Computational Linguistics",
    url = "https://aclanthology.org/2026.findings-eacl.276/",
    doi = "10.18653/v1/2026.findings-eacl.276",
    pages = "5223--5247"
}

@misc{liu2026largereasoningmodelsnot,
      title={Large Reasoning Models Are (Not Yet) Multilingual Latent Reasoners}, 
      author={Yihong Liu and Raoyuan Zhao and Hinrich Schütze and Michael A. Hedderich},
      year={2026},
      eprint={2601.02996},
      archivePrefix={arXiv},
      primaryClass={cs.CL},
      url={https://arxiv.org/abs/2601.02996}, 
}

@inproceedings{okewunmi-etal-2025-evaluating,
    title = "Evaluating Robustness of {LLM}s to Typographical Noise in {Y}or{\`u}b{\'a} {QA}",
    author = "Okewunmi, Paul  and
      James, Favour  and
      Fajemila, Oluwadunsin",
    editor = "Lignos, Constantine  and
      Abdulmumin, Idris  and
      Adelani, David",
    booktitle = "Proceedings of the Sixth Workshop on African Natural Language Processing (AfricaNLP 2025)",
    month = jul,
    year = "2025",
    address = "Vienna, Austria",
    publisher = "Association for Computational Linguistics",
    url = "https://aclanthology.org/2025.africanlp-1.29/",
    doi = "10.18653/v1/2025.africanlp-1.29",
    pages = "195--202"
}

\appendix

\begin{figure*}
    \centering
    \setlength{\belowcaptionskip}{-0.2cm}
    \includegraphics[width=0.98\textwidth]{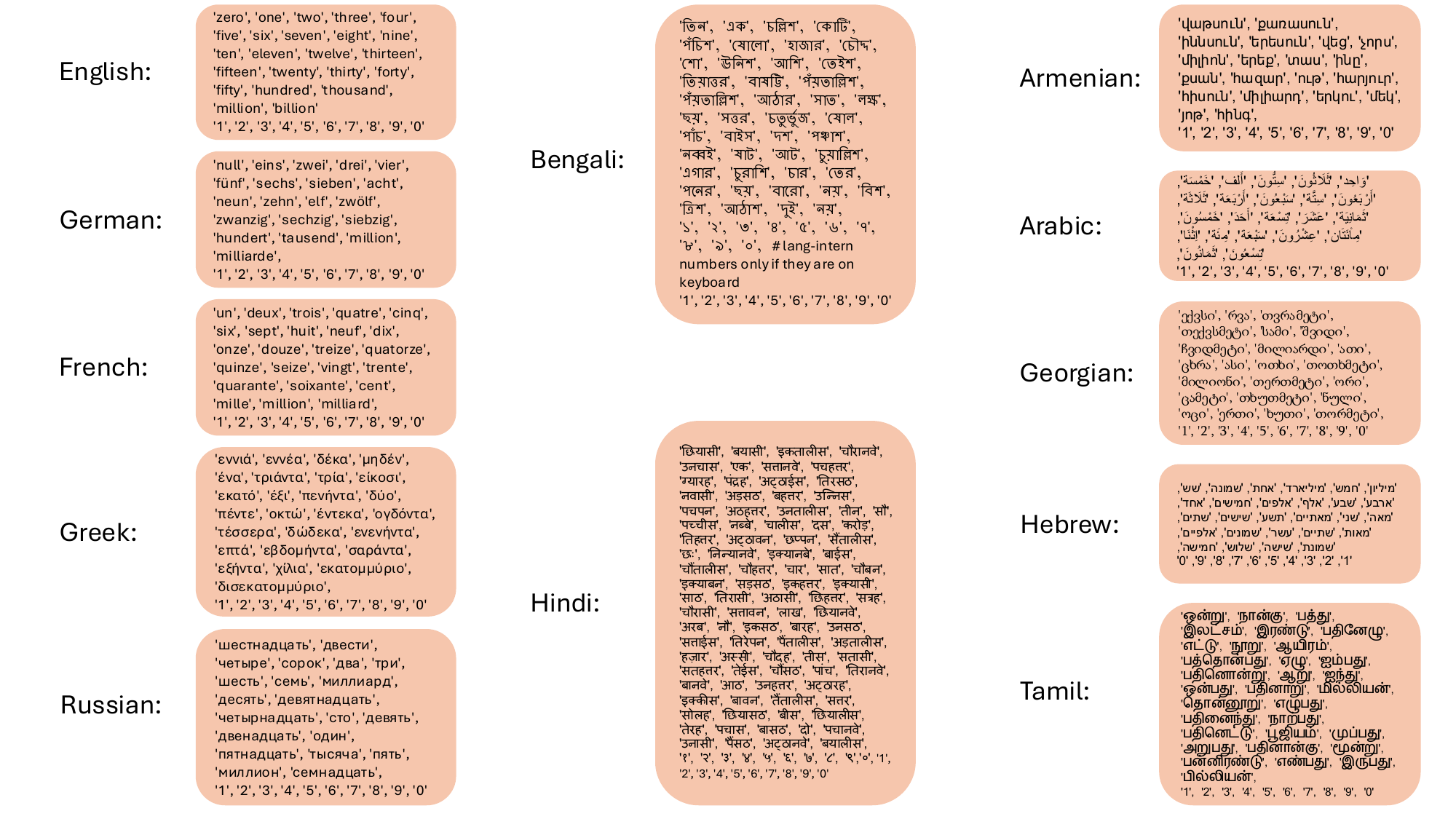}
    \caption{Ignoring String Sets across all languages that are considered in \algorithmname.}
    \label{fig:ignoring_set}
\end{figure*}

\section{Additional Details of \algorithmname}\seclabel{algorithm_detail}

\paragraph{Neighbors of Keys}
In our work, we define a character's neighbors (which are used in typo operation \emph{replace} and \emph{insert}) as the keys that are immediately adjacent to it on the same row of the keyboard -- specifically, the key to its left and the key to its right, based on the empirical findings that horizontal neighbor errors are more common \citep{macneilage1964typing}.\footnote{To reflect the real typing behavior of users of different languages, we leverage a keyboard layout database from \url{https://kbdlayout.info/}, which uses data sourced from Windows version 10.0.27729.1000}
Keys corresponding to non-alphabetic symbols or function keys (such as \emph{Enter} or \emph{Tab}) are not considered, and characters that are absent from the specified keyboard layout simply have no neighbors.
This definition provides a consistent notion of adjacency across languages and platforms, avoiding complications introduced by layout-specific variations in the placement of special characters (e.g., brackets on Windows vs. macOS).

\paragraph{Ignoring String Sets}
As described in \secref{algorithm_design}, we exclude several words from being selected as the candidates to insert typos.
These words are mainly commonly used numerical expressions across languages.
We exclude them since they are typically critical for downstream understanding.
The set for each language is visualized in Figure~\ref{fig:ignoring_set}.

\paragraph{Sampling}
In \algorithmname, there are three places where we introduce randomness by sampling: \emph{word sampling}, \emph{position sampling}, and \emph{typo operation sampling}.
We introduce the details of each sampling process in the following.

\begin{itemize}
    \item \textbf{\emph{Word Sampling}:}
    We assign each word a sampling probability proportional to the square root of its length: ${\sqrt{|w|}}$.
    We then normalize the probability by dividing the sum over all words: ${\sum_w{\sqrt{|w|}}}$.
    This probability reflects the empirically observed tendency for longer words to attract more typos.

    \item \textbf{\emph{Position Sampling}:}
    When selecting a specific character position within a word to insert or modify a character, we consider position-dependent weights: the first character is assigned a weight of 0 and is never selected; the second character receives a weight of 0.1; the final character receives 0.2; and all intermediate positions are linearly interpolated between these values. 
    This empirical setup is based on findings from \citet{lisbach2013linguistic} that word-initial errors are rare.
    The probability of each position is then its normalized weight.

    \item \textbf{\emph{Typo Operation Sampling}:} \emph{Insertion} operations are sampled with a probability of 15.25\%, while \emph{replacement}, \emph{deletion}, and \emph{transposition} are each assigned a probability of 28.25\%. 
    This skewed distribution reflects empirical observations that insertion errors are less common than the other three error types according to the findings from \citet{baba-suzuki-2012-spelling}.
\end{itemize}

\paragraph{Validating Operation}
In our implementation, instead of sampling all candidate words at once, we iteratively select a word at a time and insert a typo into it, guaranteeing an adequate number of typos according to the user specification (typo rate).
Therefore, when inserting a typo into a word, we perform a validity check before applying each operation. 
Because errors are introduced iteratively, unconstrained edits could yield implausible outcomes -- for example, replacing the initial \emph{w} in \emph{word} with \emph{e} to obtain \emph{eord}, and then replacing \emph{e} back with \emph{w}, which would be counted as two errors despite leaving the word unchanged. 
The validity check prevents such contradictions and filters out unlikely multi-step substitutions, ensuring that the final set of typos is consistent with natural typing patterns.

\paragraph{Special Cases and Strategies}
We handle several edge cases to keep the typographical errors meaningful and realistic:
\begin{itemize}
    \item If the selected word is only \emph{one character long}, or if it contains an item of the predefined Ignoring String Sets, another word will be selected.
    \item If the \emph{transpose} operation is selected and the word is not the final word in the sentence, a whitespace character is appended to the end of the word. This is done to facilitate the detection of cross-hand key pairs, as previously described, unless the whitespace has already been appended in an earlier iteration.
    \item  If the typo operation is deemed invalid or if the modified word is identical to the original (indicating no actual change occurred), another typo function is selected for the same word.
\end{itemize}

\section{Details of Human Evaluation}\seclabel{human_eva}

\begin{figure}
    \centering
    \includegraphics[width=0.45\textwidth]{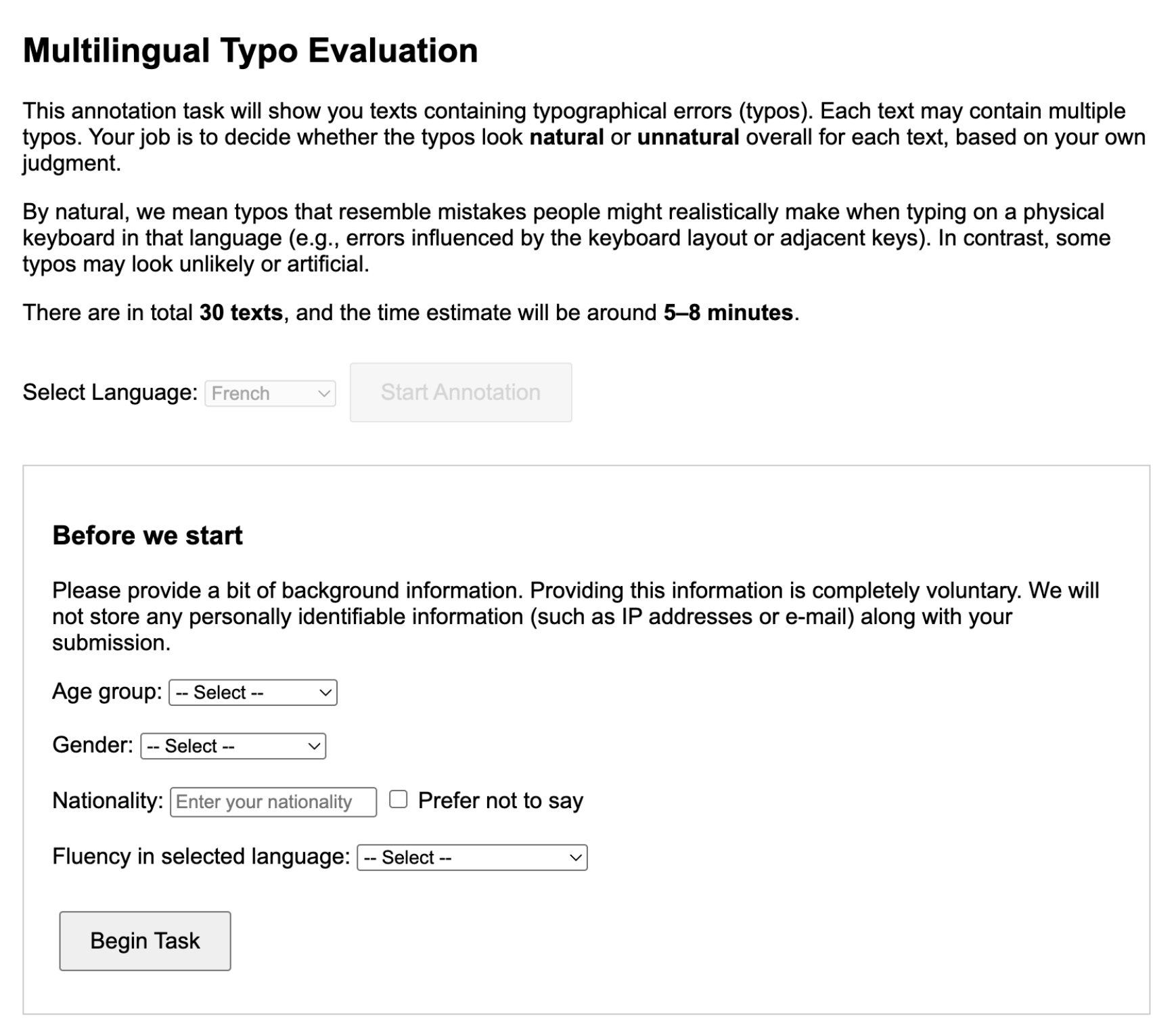}
    \caption{Screenshot of the annotation interface.}
    \label{fig:interface1}
\end{figure}

\begin{figure}
    \centering
    \includegraphics[width=0.45\textwidth]{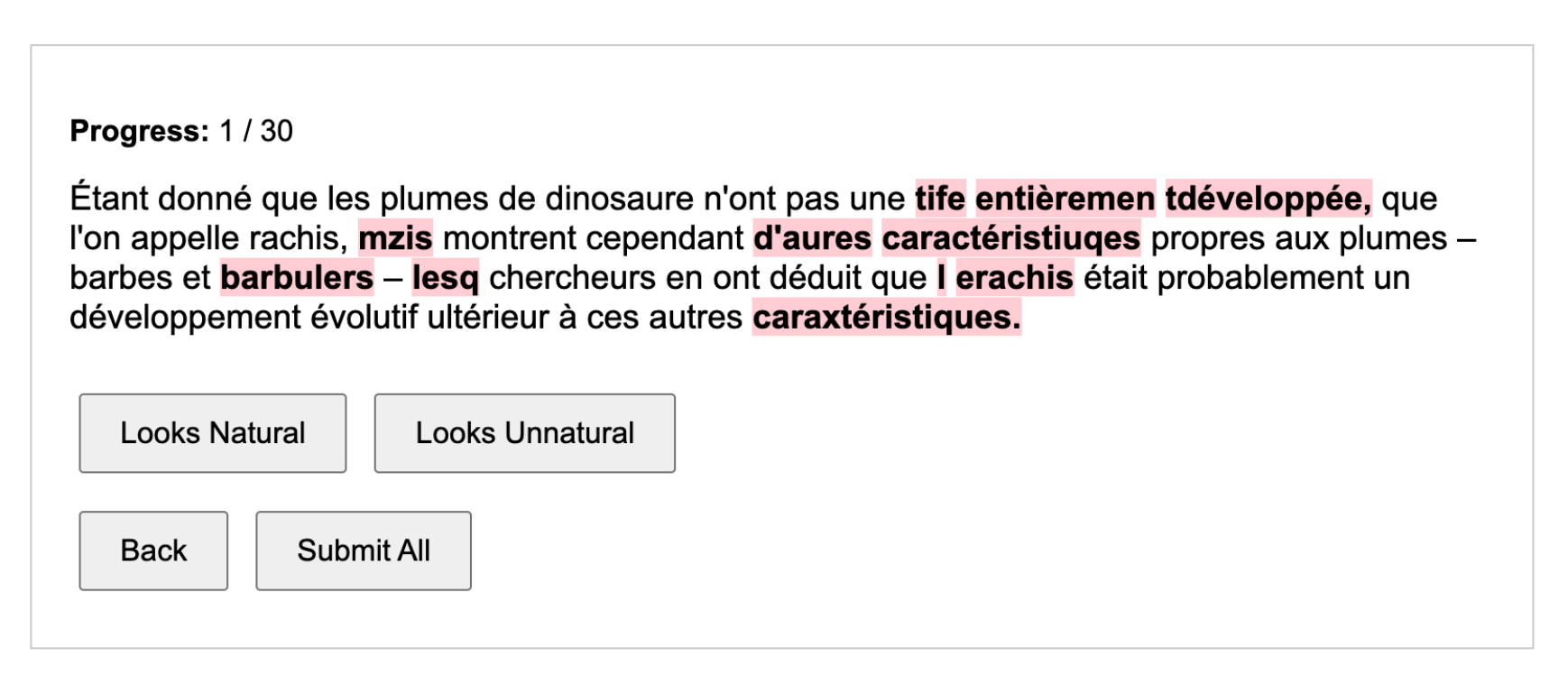}
    \caption{Example of annotating one sentence.}
    \label{fig:interface2}
\end{figure}

\paragraph{Overview}  
We designed a lightweight web interface to collect judgments of typo naturalness. 
Participants first selected their evaluation language (\textbf{English}, \textbf{German}, \textbf{French}, \textbf{Greek}, \textbf{Russian}, \textbf{Arabic}, \textbf{Hindi}), filled in basic demographic information (\emph{age}, \emph{gender}, \emph{nationality}, \emph{fluency} in the selected language),\footnote{However, providing this information is completely voluntary. We did not store any personally identifiable information except for fluency (for ensuring the quality).}
, as shown in Figure~\ref{fig:interface1}
and then completed 30 annotation trials. 
Each trial displayed a single corrupted sentence from Flores200 \citep{flores200}, and participants judged whether the typos appeared \emph{natural} or \emph{unnatural}, as shown in Figure~\ref{fig:interface2}.  
At the end of the evaluation, the participant needed to provide a confidence rating (1–5) (the higher, the more confident).  

\paragraph{Task Setup.}  
In total, 30 sentences were sampled from Flores200, with 15 corrupted by \algorithmname and 15 by the random baseline (5 sentences at 10\%, 40\%, and 70\% typo rates for each system).  
Sentences were balanced for length across the two conditions.  
The order of the sentences is randomized for each participant to avoid a systematic learning effect while going through the sentences.
Participants completed the annotation in $\sim$5-8 minutes.
Participants were recruited through personal contacts ($\sim$10\%, mainly for English and German), extended through recruitment on Prolific.\footnote{\url{https://www.prolific.com/}} We compensated crowdworkers at a rate equivalent to about £6 per hour, which corresponds to roughly £1 per annotation set (30 sentences).

\begin{figure*}
    \centering
    \includegraphics[width=0.32\textwidth]{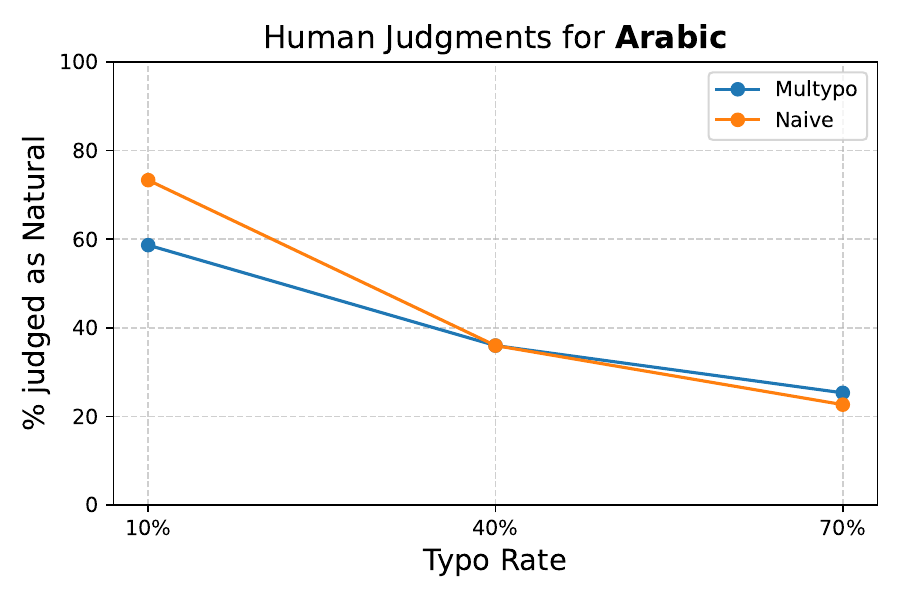}
    \includegraphics[width=0.32\textwidth]
    {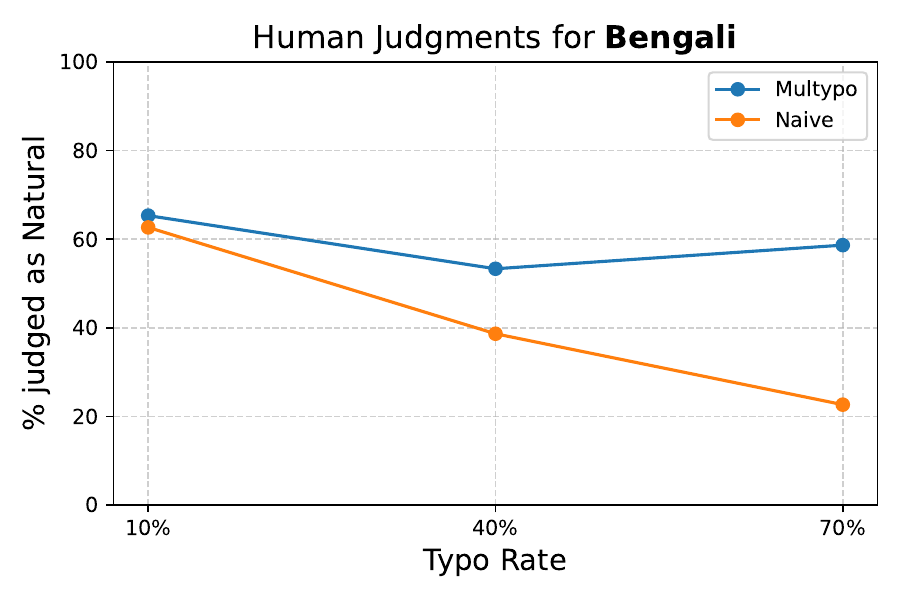}
    \includegraphics[width=0.32\textwidth]{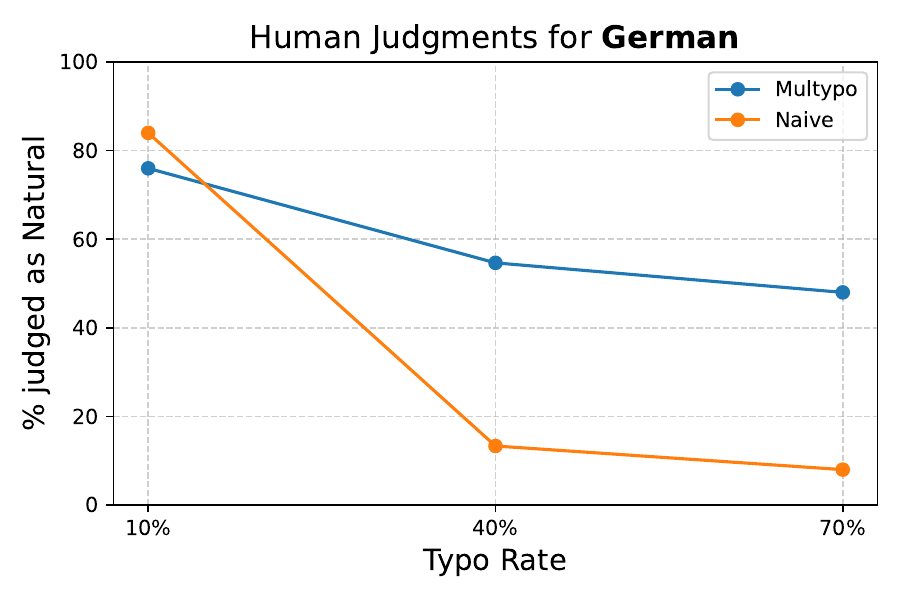}
    \includegraphics[width=0.32\textwidth]{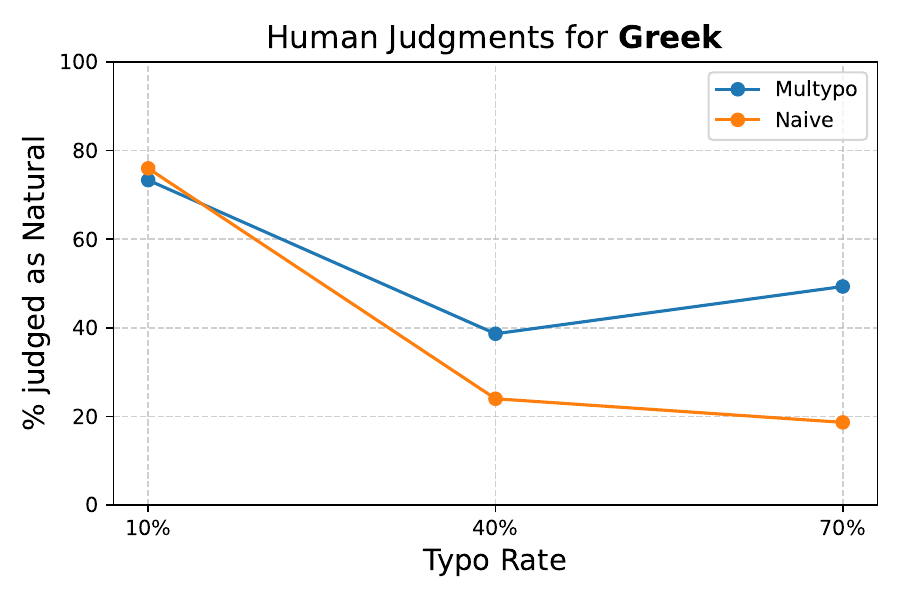}
    \includegraphics[width=0.32\textwidth]{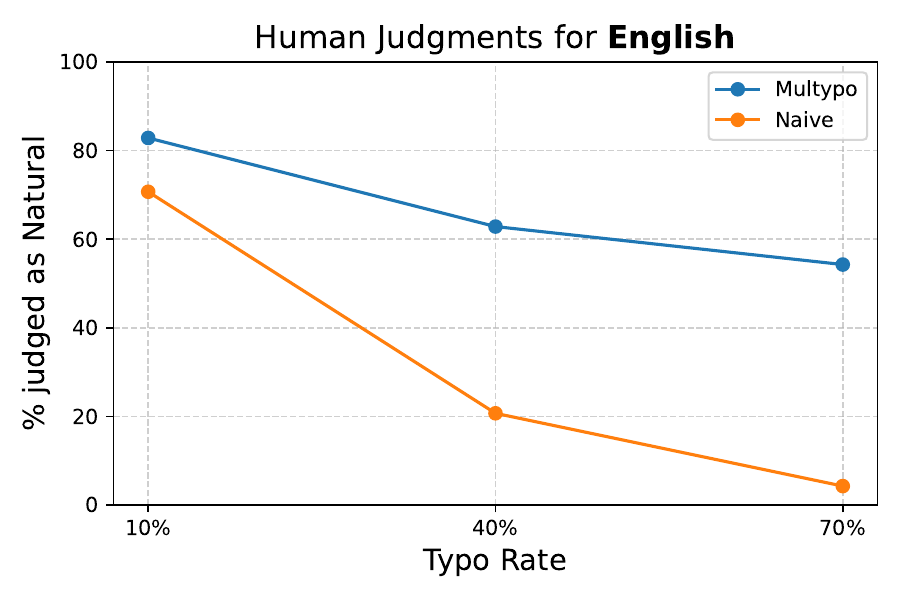}
    \includegraphics[width=0.32\textwidth]{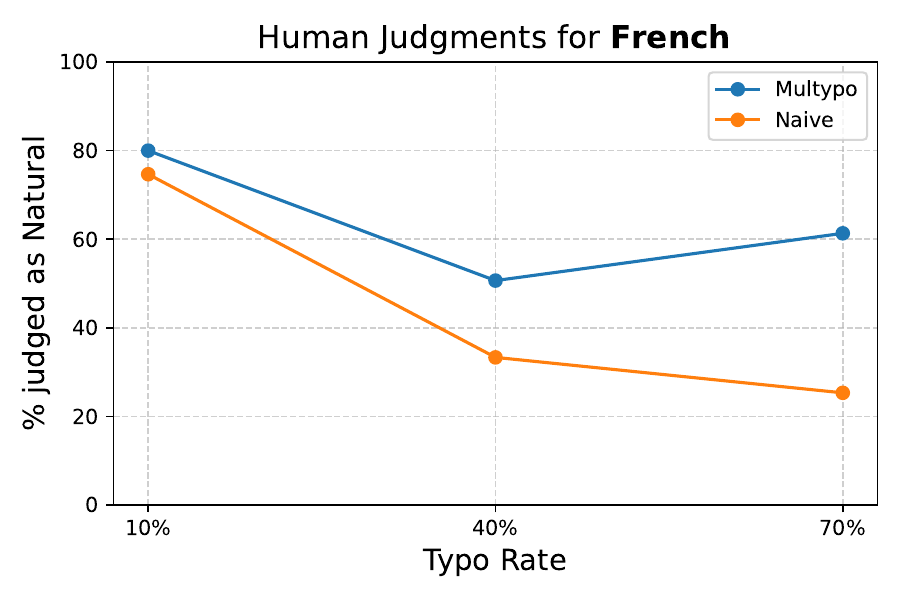}
    \includegraphics[width=0.32\textwidth]
    {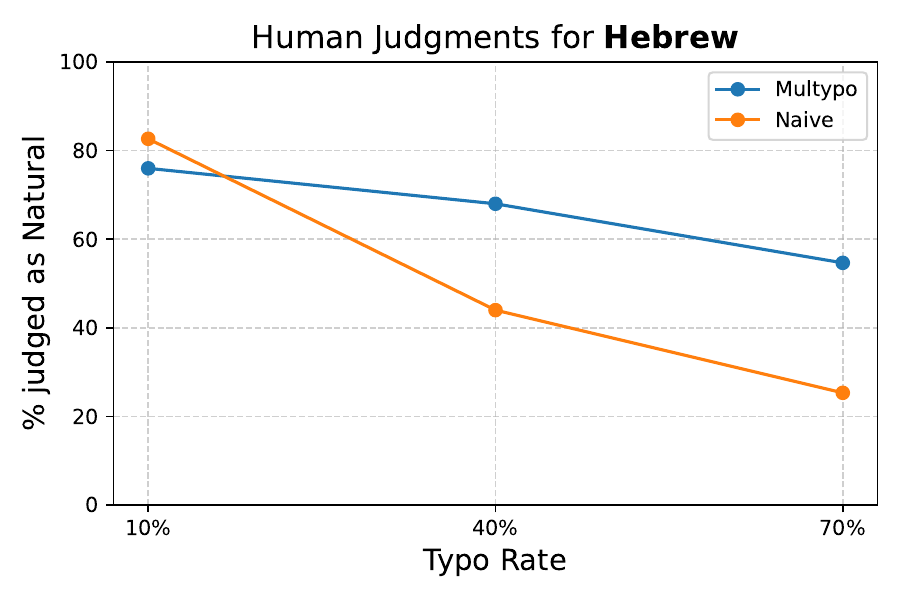}
    \includegraphics[width=0.32\textwidth]{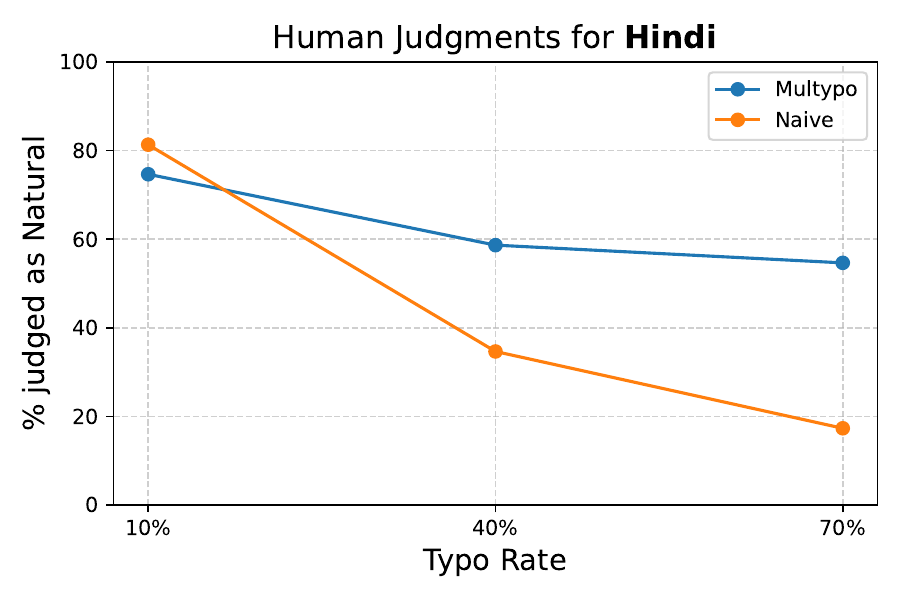}
    \includegraphics[width=0.32\textwidth]
    {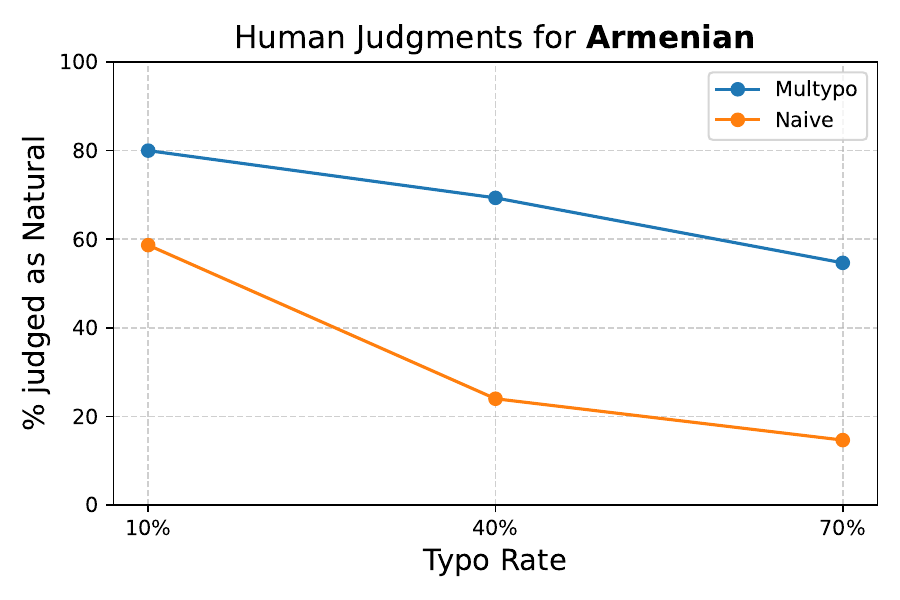}
    \includegraphics[width=0.32\textwidth]
    {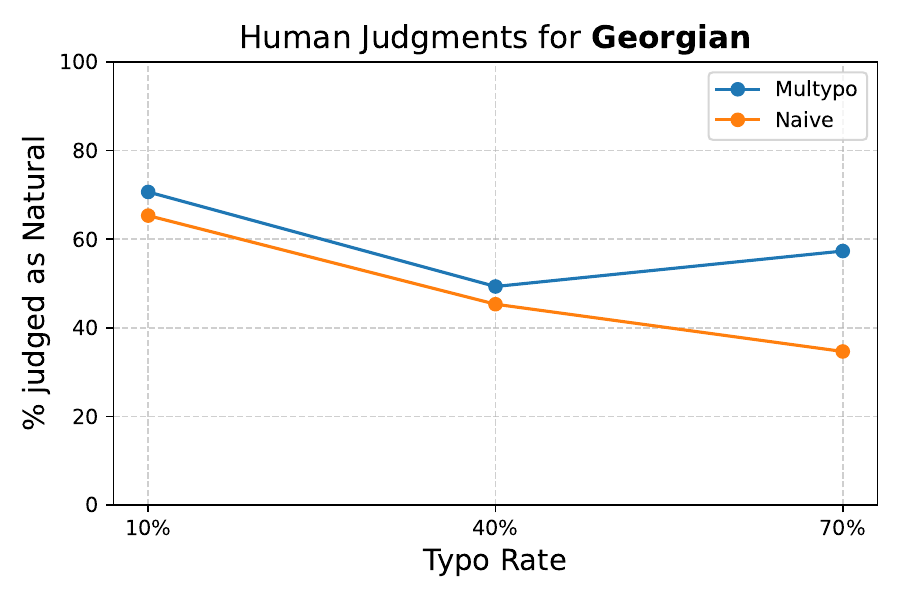}
    \includegraphics[width=0.32\textwidth]{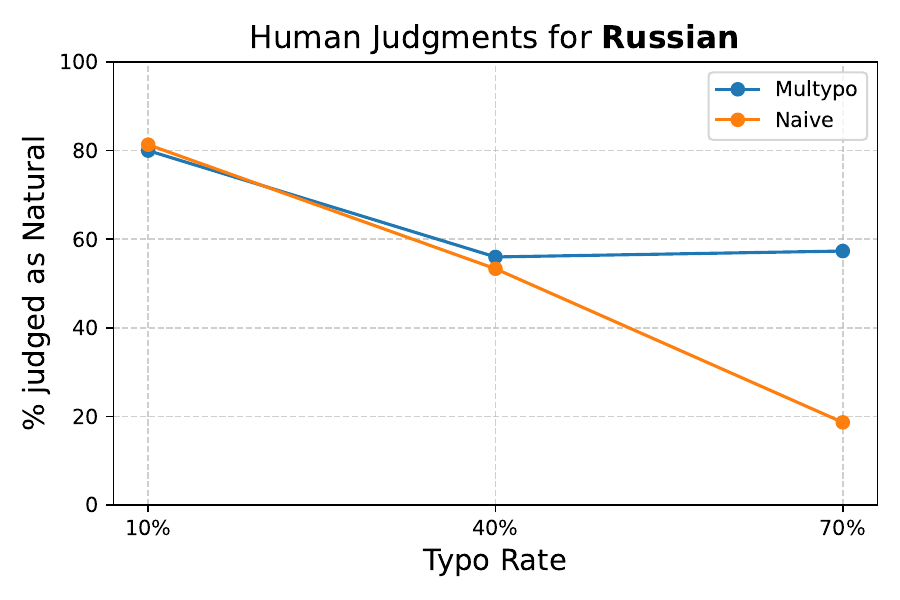}
    \includegraphics[width=0.32\textwidth]{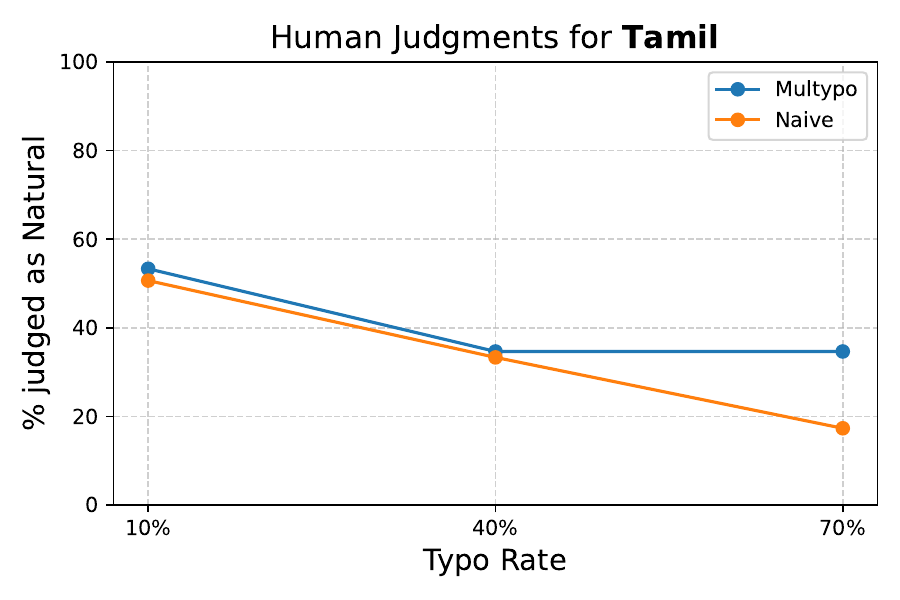}
    \caption{Human evaluation results grouped by the three considered typo rates (10\%, 40\%, 70\%) across seven languages.
    Across all languages, higher typo rates reduce perceived naturalness, yet \algorithmname consistently yields more human-like typos than the naive baseline, even under higher corruption rates.}
    \label{fig:human_eva_rate}
\end{figure*}

\paragraph{Participants.}  
We collected at least 15 valid responses per language.\footnote{For each language, we enabled Prolific’s auto-filtering feature, which excluded responses completed in under 3 minutes as likely invalid.}
Table~\ref{tab:human_eval_stats} summarizes the fluency levels and self-reported confidence across languages. 
Table~\ref{tab:annotator_age} and Table~\ref{tab:annotator_gender} summarize the distribution of age and gender, respectively.
Overall, the pool covered a balanced mix of \emph{native}, \emph{near-native}, and \emph{non-native} speakers, with most participants being native or near-native and reporting high confidence (4–5). 

\begin{table}[t]
\centering
\setlength{\belowcaptionskip}{-0.3cm}
\resizebox{\linewidth}{!}{
\begin{tabular}{lcccc}
\hline
\textbf{Language} & \textbf{Native} & \textbf{Near-native} & \textbf{Non-native} & \textbf{Avg. Confidence (1–5)} \\
\hline
Arabic    & 7 & 4 & 4 & 4.1 \\
Armenian  & 6 & 1 & 8 & 3.5 \\
Bengali   & 3 & 4 & 8 & 3.3 \\
English   & 13 & 15 & 0 & 3.9 \\
French    & 5 & 5 & 5 & 3.9 \\
Georgian  & 4 & 2 & 9 & 3.5 \\
German    & 6 & 7 & 2 & 4.0 \\
Greek     & 8 & 3 & 4 & 3.8 \\
Hebrew    & 4 & 5 & 6 & 3.5 \\
Hindi     & 8 & 2 & 5 & 4.0 \\
Russian   & 7 & 4 & 4 & 4.0 \\
Tamil     & 6 & 6 & 3 & 3.9 \\
\hline
\end{tabular}
}
\caption{Participant language fluency by language: number of native/near-native (very fluent)/non-native (basic) speakers and average self-reported confidence.}
\label{tab:human_eval_stats}
\end{table}

\begin{table}[h]
\centering
\setlength{\belowcaptionskip}{-0.2cm}
\resizebox{\linewidth}{!}{
\begin{tabular}{lcccccc}
\hline
\textbf{Language} & \textbf{18--24} & \textbf{25--34} & \textbf{35--44} & \textbf{45--54} & \textbf{55+} & \textbf{Prefer not} \\
\hline
Arabic    & 5 & 6 & 3 & 1 & -- & -- \\
Armenian  & 2 & 7 & 4 & 1 & 1 & -- \\
Bengali   & 4 & 9 & -- & 1 & -- & 1 \\
English   & 6 & 9 & 2 & 2 & 3 & 6 \\
French    & 2 & 6 & 4 & 3 & -- & -- \\
Georgian  & 3 & 9 & 2 & 1 & -- & -- \\
German    & 1 & 1 & 3 & 3 & 1 & 6 \\
Greek     & 4 & 4 & 5 & -- & 1 & 1 \\
Hebrew    & 5 & 6 & 3 & 1 & -- & -- \\
Hindi     & 3 & 9 & 2 & 1 & -- & -- \\
Russian   & 1 & 8 & 2 & 2 & 2 & -- \\
Tamil     & 5 & 7 & 1 & 2 & -- & -- \\
\hline
\end{tabular}
}
\caption{Annotator age distribution across languages. Dashes indicate that no participant reported the corresponding category.}
\label{tab:annotator_age}
\end{table}

\begin{table}[h]
\centering
\setlength{\belowcaptionskip}{-0.2cm}
\resizebox{\linewidth}{!}{
\begin{tabular}{lcccc}
\hline
\textbf{Language} & \textbf{Male} & \textbf{Female} & \textbf{Prefer not} & \textbf{Unspecified} \\
\hline
Arabic    & 6 & 9 & -- & -- \\
Armenian  & 6 & 9 & -- & -- \\
Bengali   & 8 & 6 & -- & 1 \\
English   & 9 & 13 & 6 & -- \\
French    & 8 & 5 & -- & 2 \\
Georgian  & 4 & 11 & -- & -- \\
German    & 4 & 5 & 6 & -- \\
Greek     & 8 & 6 & 1 & -- \\
Hebrew    & 7 & 7 & 1 & -- \\
Hindi     & 7 & 8 & -- & -- \\
Russian   & 6 & 8 & -- & 1 \\
Tamil     & 9 & 6 & -- & -- \\
\hline
\end{tabular}
}
\caption{Annotator gender distribution across languages. ``Unspecified'' refers to missing or null responses.}
\label{tab:annotator_gender}
\end{table}

\paragraph{Findings.}  

Extending on the results reported in \secref{user_study}, 
we also observe a clear effect of corruption level (cf. Figure~\ref{fig:human_eva_rate}): across all languages, higher typo rates lead to substantially lower ``naturalness'' judgments, aligning with the intuition that dense error patterns are less plausible as real-world human mistakes.  
Importantly, we observe that even under severe corruption (i.e., $40\%$ and $70\%$), \algorithmname maintains a naturalness advantage over the naive baseline, highlighting that modeling human-typing behavior provides a better simulation of real-world typos.  

\subsection{LLM-as-a-Judge Evaluation of \algorithmname Naturalness}\seclabel{llm_as_a_judge}

\begin{table}[t]
\centering
\footnotesize
\setlength{\belowcaptionskip}{-0.5cm}
\resizebox{\linewidth}{!}{
\begin{tabular}{lrrr}
\hline
\textbf{Language} & \textbf{Multypo (avg.)} & \textbf{Naive (avg.)} & \textbf{Significance} \\
\hline
Arabic     & 7.10  & 6.70  &     \\
Armenian   & 12.20 & 0.60  & *** \\
Bengali    & 11.40 & 4.40  & *** \\
English    & 13.50 & 7.20  & *** \\
French     & 12.40 & 5.00  & *** \\
Georgian   & 13.40 & 6.70  & *** \\
German     & 12.70 & 5.60  & *** \\
Greek      & 10.90 & 7.80  & **  \\
Hebrew     & 10.20 & 8.00  & **  \\
Hindi      & 10.20 & 3.00  & *** \\
Russian    & 10.20 & 5.60  & *** \\
Tamil      & 5.90  & 4.40  & *   \\
\hline
\end{tabular}
}
\caption{LLM-as-a-judge evaluation of typo naturalness. We report the average number of times (out of 10 runs) that a typo-corrupted sentence is classified as ``natural'' by the LLM for each method. Significance is computed using paired t-tests. Stars denote significance levels: * $p<0.05$, ** $p<0.01$, *** $p<0.001$.}
\label{tab:llm_judge}
\end{table}

To further assess the naturalness of typos generated by \algorithmname, we conduct an additional evaluation using \texttt{Gemini-2.5-Flash} as a judge, following recent work on model-based evaluation.
We adopt a protocol aligned with our human study (cf. \secref{user_study}): given a typo-corrupted sentence, the LLM is prompted to classify it as either \textit{natural} or \textit{unnatural}, using the same instructions as in human annotation. 
To better approximate multiple independent annotators, we repeat the evaluation 10 times for each sentence and aggregate the results.
Specifically, we report the average number of times (out of 10) that a sentence is classified as \textit{natural} for each typo generation method. 
We further compute paired t-tests across methods for each language.

Table~\ref{tab:llm_judge} presents the results of LLM-as-a-judge. 
We observe that \algorithmname consistently produces more natural typos than the naive baseline across almost all languages, with statistically significant improvements in the majority of cases. 
The trends are highly consistent with our human evaluation results reported in Table \ref{tab:user_study_nat} in \secref{user_study}, providing additional evidence that \algorithmname better captures realistic human typing behavior. 
While the gap is smaller for some languages (e.g., Arabic, Tamil), the overall pattern remains robust.

\section{Details of Downstream Tasks}\seclabel{downstream_detail}

\subsection{Dataset Statistics}

\paragraph{Language Coverage}

\algorithmname supports 12 languages at the current stage.
Each downstream task covers a slightly different set of languages, and therefore, we only evaluate on languages that are supported by \algorithmname for each dataset.
Table~\ref{tab:datasets-languages} presents the languages supported in each dataset.

\paragraph{Instance Selection}

To ensure a fair and balanced evaluation, we cap each dataset at a maximum of 1,000 instances across languages where possible. 
Belebele contains roughly 900 instances by design and requires no further reduction. 
For Flores200, we selected 500 instances per translating direction (translating into and from English), yielding 1,000 prompts in total. 
MGSM and its Arabic and Hindi adaptations are limited to 250 examples each to maintain consistency. 
For XNLI and MMMLU, we subsample 1,000 instances while preserving original label (XNLI) and subject-area (MMMLU) distributions.
The sampling is consistently applied in each parallel dataset to ensure comparability across languages.

\begin{table*}
\centering
\footnotesize
\setlength{\tabcolsep}{3pt}
\resizebox{\textwidth}{!}{
\begin{tabular}{lcccccccccccc}
\hline
{Dataset} & {ara\_Arab} & {ben\_Beng} & {deu\_Latn} & {ell\_Grek} & {eng\_Latn} & {fra\_Latn} & {heb\_Hebr} & {hin\_Deva} & {hye\_Armn} & {kat\_Geor} & {rus\_Cyrl} & {tam\_Taml} \\ \hline
XNLI       & x &   & x & x & x & x &   & x &   &   & x &   \\ 
Belebele   & x & x & x & x & x & x & x & x & x & x & x & x \\ 
MMMLU      & x & x & x &   & x & x &   & x &   &   &   &   \\ 
MGSM       & x & x & x &   & x & x &   & x &   &   & x &   \\ 
FLORES200  & x & x & x & x & x & x & x & x & x & x & x & x \\ \bottomrule
\end{tabular}
}
\caption{Supported languages for each dataset in our evaluation.}
\label{tab:datasets-languages}
\end{table*}

\subsection{Prompt Templates}

For each dataset, prompts are constructed in a standardized format to ensure consistency across experiments. 
If typographical errors are injected, only the instances of the dataset, denoted in curly brackets \{\} (\{\texttt{language}\} in Flores200 is an exception), are affected. All other components of the prompt remain unchanged.
The prompt templates are shown as follows.

\begin{PromptBox}{XNLI}
\ttfamily\footnotesize
Classify the relationship between the premise and hypothesis as
(0) Entailment, (1) Neutral, or (2) Contradiction.\\
\textbf{Premise:} \{\texttt{premise}\}\\
\textbf{Hypothesis:} \{\texttt{hypothesis}\}\\
\textbf{Label:}\,
\end{PromptBox}

\begin{PromptBox}{Belebele}
\ttfamily\footnotesize
Given the following passage, query, and answer choices, output the letter of the correct answer.\\
\#\#\#\\
\textbf{Passage:}\\
\{\texttt{flores\_passage}\}\\
\#\#\#\\
\textbf{Query:}\\
\{\texttt{question}\}\\
\#\#\#\\
\textbf{Choices:}\\
(A) \{\texttt{mc\_answer1}\}\\
(B) \{\texttt{mc\_answer2}\}\\
(C) \{\texttt{mc\_answer3}\}\\
(D) \{\texttt{mc\_answer4}\}\\
\#\#\#\\
\textbf{Answer:}
\end{PromptBox}

\begin{PromptBox}{MMMLU}
\ttfamily\footnotesize
The following are multiple choice questions (with answers) about \{\texttt{Subject}\}.\\
\{\texttt{Question}\}\\
(A) \{\texttt{A}\}\; (B) \{\texttt{B}\}\; (C) \{\texttt{C}\}\; (D) \{\texttt{D}\}\\
\textbf{Answer:}
\end{PromptBox}

\begin{PromptBox}{MGSM}
\ttfamily\footnotesize
Question: \{\texttt{question}\} \; Let's think step by step.\\
\textbf{Step-by-Step Answer:}
\end{PromptBox}

\begin{PromptBox}{Flores200: \{\texttt{language}\} $\rightarrow$ English}
\ttfamily\footnotesize
Translate the following sentence from \{\texttt{language}\} to English.\\
\{\texttt{language}\}: \texttt{<BOS>}\{\texttt{sentence}\}\texttt{<EOS>}\\
English: \texttt{<BOS>}
\end{PromptBox}

\begin{PromptBox}{Flores200: English $\rightarrow$ \{\texttt{language}\}}
\ttfamily\footnotesize
Translate the following sentence from English to \{\texttt{language}\}.\\
English: \texttt{<BOS>}\{\texttt{sentence}\}\texttt{<EOS>}\\
\{\texttt{language}\}: \texttt{<BOS>}
\end{PromptBox}

\paragraph{Few-Shot Setup}
For each shot setting, we sample a fixed set of support examples per language and introduce typographical noise according to the specified corruption level. 
To preserve a consistent supervision signal, the correct answers in the exemplars are left unaltered, allowing the model to learn correct associations even under noisy contexts.
The same exemplars are used across all evaluations to ensure comparability. 
Whenever possible, examples are drawn from the training or development splits; otherwise, the first instances from the test split are selected. 
The remaining data points serve as independent queries during evaluation.

\section{Performance Comparison with Random Typo Baseline}\seclabel{comparison_baseline}

\begin{figure*}
    \centering
    \setlength{\abovecaptionskip}{-0.05cm}
    \setlength{\belowcaptionskip}{-0.4cm}
    \includegraphics[width=0.98\textwidth]{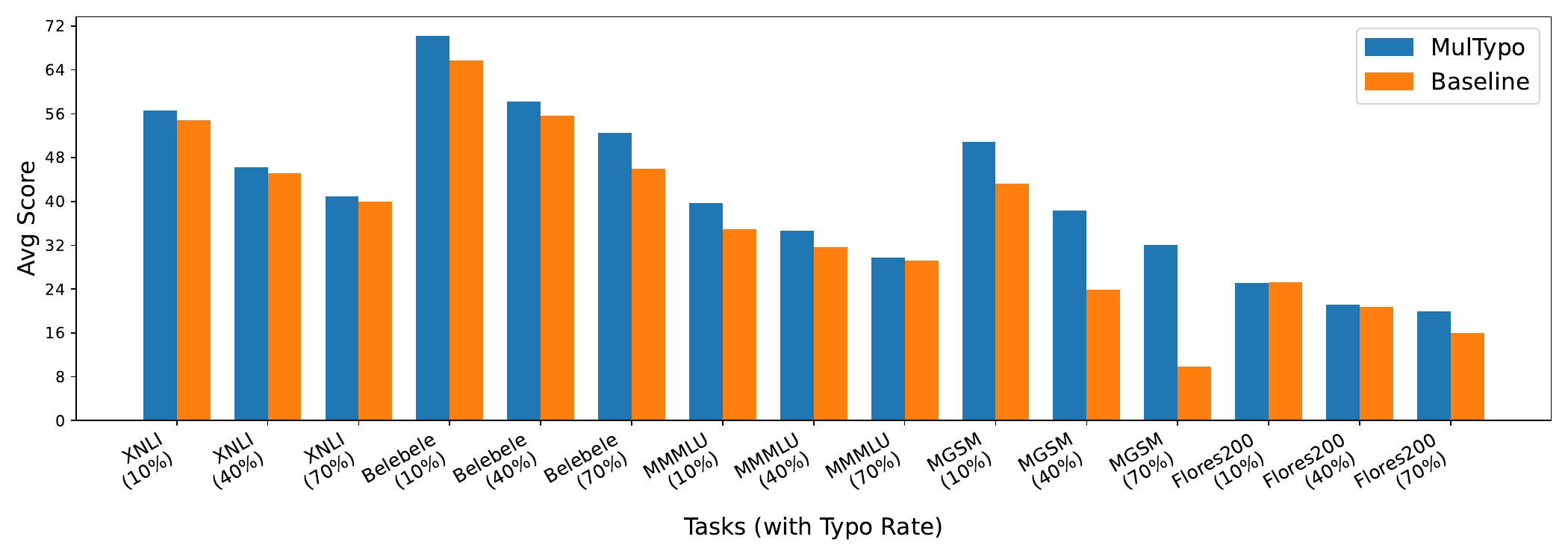}
    \caption{Performance comparison of \textsc{MulTypo} and random baseline under different typo rates (10\%, 40\%, 70\%) in the 3-shot setting. Bars show average performance across all languages for each task.}
    \label{fig:baseline_comparison}
\end{figure*}


We examine performance differences when input text is perturbed by \textsc{MulTypo} -- which simulates human typing behavior -- versus a \emph{random typo baseline} that applies the same four operations described in \secref{algorithm_design} but disregards keyboard layout constraints.
For this evaluation, we use \texttt{Qwen3-4B} (instruction-tuned) across all five tasks.

\begin{table}[t]
\centering
\begin{tabular}{llll}
\hline
Task & 10\% & 40\% & 70\% \\ \hline
XNLI & 0.665 & 0.621 & 0.768 \\
Belebele & 0.001*** & 0.029* & 0.000*** \\
MMMLU & 0.133 & 0.170 & 0.168 \\
MGSM & 0.002** & 0.001** & 0.001*** \\
Flores200 & 0.213 & 0.071 & 0.000*** \\
\hline
\end{tabular}
\caption{Significance of performance differences between \textsc{MulTypo} and \emph{random baseline} under 3-shot setup. Each cell shows the $p$-value with significance stars (* $p<0.05$, ** $p<0.01$, *** $p<0.001$).}
\label{tab:significance}
\end{table}

As shown in Figure~\ref{fig:baseline_comparison}, performance under random perturbations is consistently lower than with \textsc{MulTypo}.
This suggests that models are more robust to \textsc{MulTypo} typos, likely because they better approximate realistic human typing errors, some of which may already be represented in pretraining corpora.
Table~\ref{tab:significance} further confirms these trends:  
(1) For natural language understanding tasks (e.g., XNLI), the performance gap is small and not statistically significant.  
(2) For generation tasks -- particularly those requiring reasoning -- the random baseline leads to significantly larger degradation compared to \textsc{MulTypo}.

\section{Complementary Results on Translation}\seclabel{complementary_results}

\begin{figure*}
    \centering
    \setlength{\abovecaptionskip}{-0.05cm}
    \setlength{\belowcaptionskip}{-0.4cm}
    \includegraphics[width=0.98\textwidth]{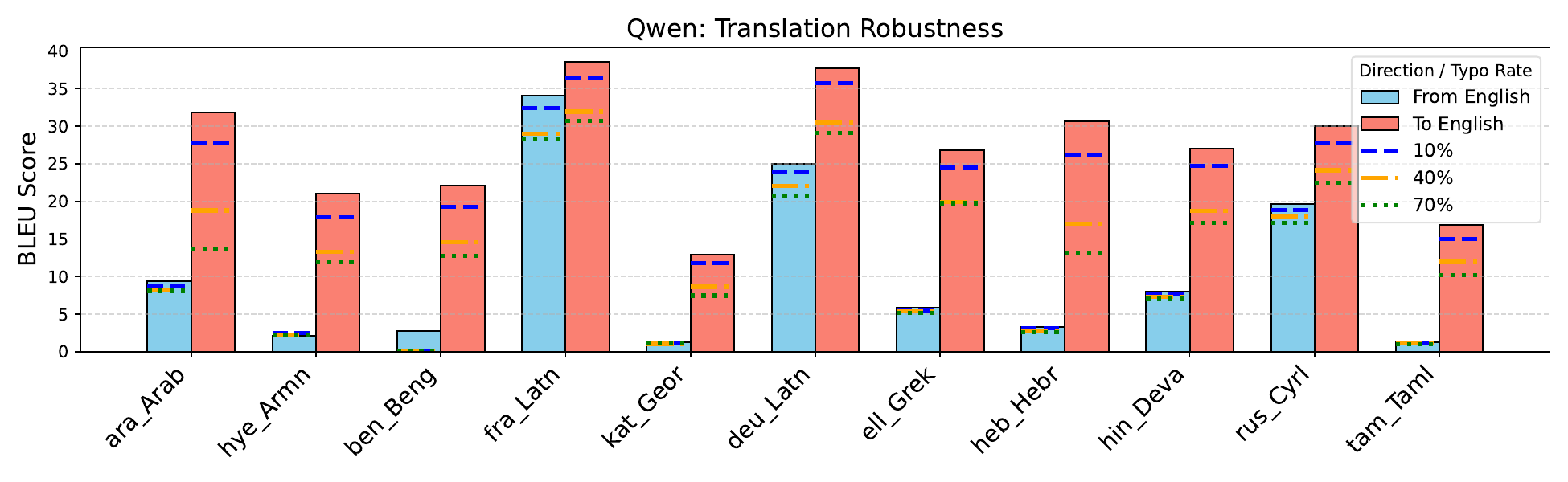}
    \caption{Robustness of \textbf{Qwen} models on \textbf{Flores200} under different levels of typographical noise.
    Translation from English seems to be more robust compared to translation to English.}
    \label{fig:flores_qwen}
\end{figure*}

\begin{figure*}
    \centering
    \setlength{\abovecaptionskip}{-0.05cm}
    \setlength{\belowcaptionskip}{-0.4cm}
    \includegraphics[width=0.98\textwidth]{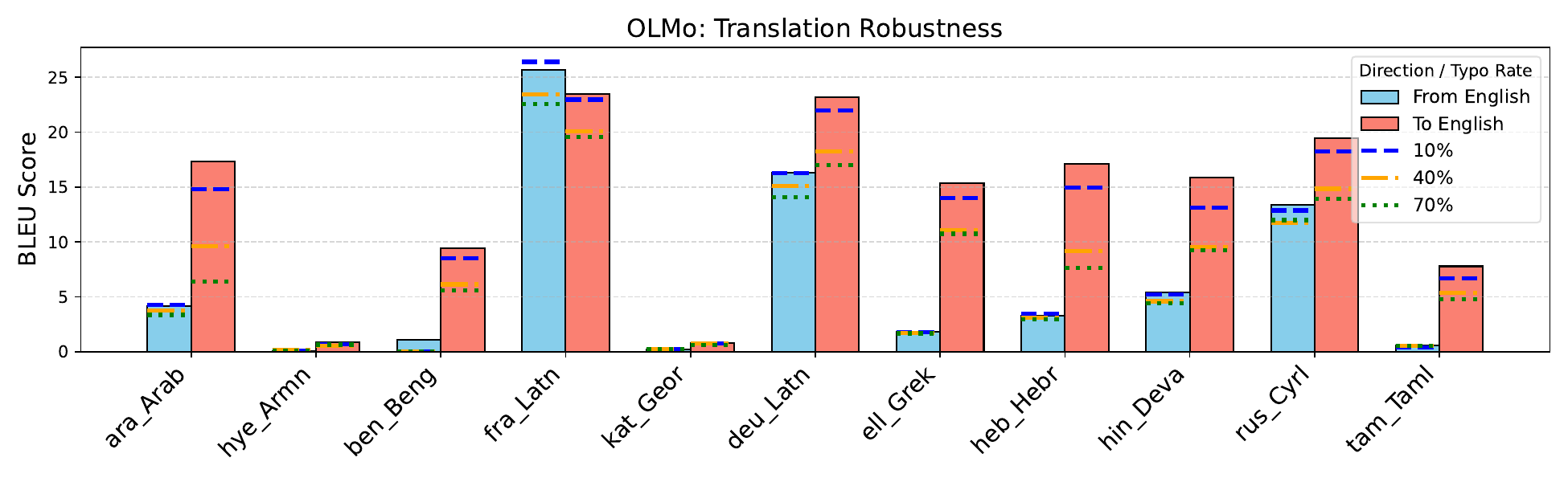}
    \caption{Robustness of \textbf{OLMo} models on \textbf{Flores200} under different levels of typographical noise.
    Translation from English seems to be more robust compared to translation to English.}
    \label{fig:flores_olmo}
\end{figure*}

We also compare the performance of \textbf{Qwen} and \textbf{OLMo} models when translating \emph{from English} vs. \emph{into English}, as shown in Figure~\ref{fig:flores_qwen} and Figure~\ref{fig:flores_olmo}.
In general, we observe the same trend as in Figure~\ref{fig:flores_gemma}.
That is, translations \emph{from} English are more robust than those \emph{into} English.
This trend is typically noticeable when involving low-resource languages that are written in non-Latin scripts, such as hye\_Armn and kat\_Geor.
To sum up, these findings further support our claim that typos in lower-resource input languages might severely impair the understanding and, therefore, result in bad translation quality.

\section{Experimental Environment and
Hyperparameters}\seclabel{environment}

All experiments are conducted on NVIDIA RTX
A6000 GPUs.
We use vLLM to process the prompts and obtain the response for each prompt.\footnote{\url{https://docs.vllm.ai/en/v0.7.3/}}
The default sampling parameters (top-$k$, top-$p$, etc.) of vLLM are used.
We set the different maximum generation tokens for each dataset: 5 for XNLI, 100 for Belebele, 5 for MMMLU, 200 for MGSM, and 100 for Flores200, based on preliminary experimental results.

\end{document}